\documentclass[letterpaper, journal]{IEEEtran} 
\IEEEoverridecommandlockouts

\usepackage{cite}

\def\BibTeX{{\rm B\kern-.05em{\sc i\kern-.025em b}\kern-.08em
    T\kern-.1667em\lower.7ex\hbox{E}\kern-.125emX}}
    
  \usepackage{amsmath,amssymb,amsfonts,amsthm} 
\usepackage{mathtools}
\usepackage{commath}
\usepackage{algorithmic}
\usepackage[ruled,vlined]{algorithm2e}
\usepackage{graphicx}
\usepackage{textcomp}
\usepackage[table]{xcolor}

\usepackage{framed}
\definecolor{shadecolor}{rgb}{.9,.9,.9}
\usepackage{tcolorbox}
\usepackage{siunitx}
\usepackage{textgreek}
\usepackage{url}
\usepackage{booktabs, tabularx}
\usepackage{multirow}
\usepackage{stackengine}
\usepackage[caption=false, font=footnotesize]{subfig}

\usepackage{tikz}
\usetikzlibrary{quotes,angles}
\usetikzlibrary{calc}
\usetikzlibrary{decorations.pathmorphing}

\usepackage{soul}
    \usepackage{todonotes}
\usepackage{tablefootnote}
\usepackage[capitalize]{cleveref}
\crefrangeformat{equation}{#3(#1)#4--#5(#2)#6}

\title{\LARGE\bf Collective Control for Arbitrary Configurations of Docked Modboats 
}

\author{Gedaliah Knizhnik and Mark Yim
\thanks{The authors are with the GRASP Laboratory, University of Pensylvannia, Philadelphia, PA 19104. 
        {\tt\footnotesize knizhnik@seas.upenn.edu}}%
}

\DeclareMathOperator{\sign}{sgn}

\DeclareMathOperator{\rank}{rank}

\newtheorem{definition}{Definition}
\newtheorem{assumption}{Assumption}
\newtheorem{theorem}{Theorem}

\begin{document}
\bstctlcite{MyBSTcontrol} 

\maketitle

\begin{abstract}
The Modboat is a low-cost, underactuated, modular robot capable of surface swimming, docking to other modules, and undocking from them using only a single motor and two passive flippers. Undocking is achieved by causing intentional self-collision between the tails of neighboring modules in certain configurations; this becomes a challenge, however, when collective swimming as one connected component is desirable. In this work, we develop a centralized control strategy to allow \textit{arbitrary} configurations of Modboats to swim as a single steerable vehicle and guarantee no accidental undocking. We also present a simplified model for hydrodynamic interactions between boats in a configuration that is tractable for real-time control. We experimentally demonstrate that our controller performs well, is consistent for configurations of various sizes and shapes, and can control both surge velocity and yaw angle simultaneously. Controllability is maintained while swimming, but pure yaw control causes lateral movement that cannot be counteracted by the presented framework.
\end{abstract}



\section{Introduction} \label{sec:intro}

Aquatic systems that can dock, undock and reconfigure are of interest to researchers and industry; they have  potential to facilitate ocean research and infrastructure by providing mobile platforms to land helicopters or drones, building bridges for larger vehicles\cite{Paulos2015}, or forming ocean-going manipulators. They can adapt to changing flow conditions or take precise measurements at small spatial scales. Such applications can  be accomplished by systems composed of large numbers of modules, but conventional wisdom has been that the individual modules must be capable of holonomic motion~\cite{Paulos2015, OHara2014, Wang2018DesignVehicle, Wang2020RoboatEnvironments}, which makes them expensive and limits their number. Allowing the individual modules to be under-actuated would  reduce their cost and allow scaling such systems to be more efficient.

In a docked aquatic system, however, challenges exist even when the individual modules are capable of holonomic motion. These challenges include robust docking and undocking~\cite{OHara2014, Mateos2019AutonomousBoats}, assembly and disassembly of floating structures from  individual modules~\cite{Seo2013AssemblyRobots, Seo2016AssemblyModules} and from substructures, minimizing disturbances and forces within the configuration~\cite{OHara2014, Paulos2015}, achieving consensus without a centralized controller~\cite{Wang2020DistributedVessels}, and finding the optimal distribution of effort across the structure~\cite{Kayacan2019Learning-basedRoboats, Park2019CoordinatedApproach}. If we seek to reduce the cost and complexity of the modules by reducing their overall mobility, these problems become much harder.

Most non-docking aquatic robots are built in either a thruster-rudder or a differential thrust arrangement~\cite{Liu2016UnmannedChallenges}, and the ability of such systems to move competently on their own has been widely explored in the literature. In either arrangement the ability of an individual module to thrust is restricted to a linear direction in a rigid docked configuration, with the ability to yaw greatly reduced due to the relative scaling of thrust and inertia that grows unfavorably for modular systems~\cite{Gabrich2020ModQuad-DoF:Quadrotors}. We can therefore consider modules that are only capable of thrust along a single axis as a reasonable model of a general aquatic surface robot, and must find a non-trivial thrust distribution to allow such modules to collaborate when docked.

A further restriction is added when considering more unique individual modules. Modboats, introduced by the authors in prior work~\cite{modboatsOnline}, utilize a unique combination of a single motor and passive flippers for propulsion and steering (originally introduced by Refael and Degani~\cite{Refael2015},\cite{Refael2018}), and have been shown to be capable of complex motions when swimming individually~\cite{Knizhnik2020a, Knizhnik2021a}. Modboats are capable of docking in a rectangular lattice through passive magnetic docks and use a tail rigidly connected to their actuating body to undock without additional actuators~\cite{Knizhnik2021}, reducing cost, mass and complexity. However, this tail --- while enabling undocking --- introduces a significant restriction on allowable thrusts and motions when collective motion is desirable. Naive application of control strategies, without consideration of this restriction, would immediately generate undesirable undocking behavior, and the docked structure would quickly disintegrate. This motivates us --- as will be presented in Sec.~\ref{sec:selfCollisionNP} --- to consider a configuration of modules that can each thrust along a single axis where all these axes are \textit{aligned}.

Moreover, a significant issue to consider when controlling any aquatic robots swimming in close proximity, and especially when they are docked together, is the effect of hydrodynamic interaction between individual actuators. While this effect certainly exists for conventional propeller-craft\cite{Tuck1974HydrodynamicShips}, it is a significant and  complex phenomenon for tail or flipper driven swimmers, such as the Modboat. Significant work in the literature has been dedicated to exploring the effects of fluid interactions, vortices, and formation on fish swimming~\cite{Weihs1973HydromechanicsSchooling,Ashraf2017SimpleSchooling,Maertens2017OptimalSwimmers,Ramananarivo2016FlowFlight,Khalid2018OnConfiguration, Becker2015HydrodynamicSwimmers, Liao2003FishActivity,Li2017NumericalModel,Li2020VortexFish,Li2021UsingFish}. At the very least, we must acknowledge that actuator dynamics must be adjusted when applied to configurations of coordinating swimmers. 

The contribution of this work, therefore, is to develop a relatively simple centralized control strategy that can be applied to \textbf{arbitrary} rectangular configurations of docked aquatic swimmers, each capable of thrusting along \textbf{a single axis},  \textbf{aligned} with each other. We show how such a strategy can be implemented using the Modboat while \textbf{guaranteeing} that modules \textbf{will not unintentionally undock} during collective swimming, and demonstrate that a \textbf{relatively simple model of hydrodynamic interaction} provides \textbf{strong performance} when tracking yaw and surge velocity. 

This work is an extension of our prior work~\cite{Knizhnik2022AmplitudeModboats}, in which such a strategy was shown for parallel configurations only. This work extends the approach to configurations of arbitrary shape, extends the collision-free guarantee, and incorporates hydrodynamic modeling that was absent in the original work.

The rest of this work is organized as follows. Sec.~\ref{sec:dynamics} presents the dynamic model for a docked configuration and our strategy for controlling it. Sec.~\ref{sec:waveform} describes how the Modboat can be driven to match the model requirements when docked, and Sec.~\ref{sec:hydrodynamics} discusses our model of hydrodynamic interactions when the modules are driven this way. Sec.~\ref{sec:selfCollisionNP} proves that unintentional collisions are impossible under the designed strategy. Finally, Sec.~\ref{sec:experiments} presents experimental verification of controller performance, which is discussed in Sec.~\ref{sec:discussion}.


\section{Dynamics} \label{sec:dynamics}

\begin{figure}[t]
    \centering
    \includegraphics[page=2, width=\linewidth]{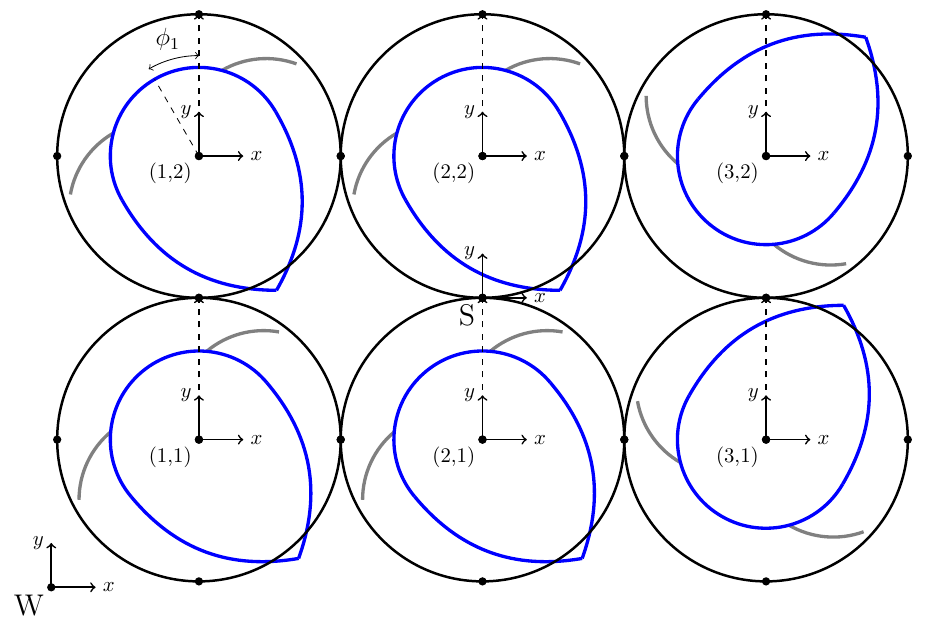}
    \caption{configuration of six docked Modboats; for each boat, the top body is shown in black, while the bottom body/tail is in blue and flippers are in gray. Magnetic docking points are shown at the cardinal points of each boat, and the motor angle $\phi_i$ is the angle of the bottom body \textit{relative} to the top body for boat $i$. Frame $W$ is the fixed world frame, and frame $S$ is the body-fixed frame at the COM, which coincides with the body-fixed frame of boat $2$. The individual boat frames are aligned with the top bodies.}
    \label{fig:diagramNP}
\end{figure}

Consider a set of swimming robotic modules arranged on a rectangular lattice, as shown in Fig.~\ref{fig:diagramNP}. Each module is capable of producing both positive and negative thrust along its own body-fixed $y$ axis on average over a period of length $T$, and we assume that all the $y$ axes are aligned with a body-fixed frame $S$ at the center of mass (COM) of the configuration. For the sake of clarity in this work we will use both 2-dimensional and linear numbering for the modules in the configuration where appropriate; a tuple of subscripts indicates a 2-dimensional numbering, and a single subscript indicates linear numbering. Any suitable mapping between the two can be used for conversions.

As long as the number of modules $N \geq 2$ and the configuration is at least two modules wide, the structure is controllable under these assumptions\footnote{As a proof, consider the equations of motion in~\eqref{eq:eom-mat1}, which is a linear equation in the form $P\vec{f} = \vec{b}$. Because $\rank{(\vec{b})}=2$, the map $P$ is \textit{surjective} as long a $\rank{P} \geq 2$. This is true as long as the configuration is at least two modules wide, meaning we have at least two unique values for $x_i$. This means any combination of acceleration and yaw torque (within actuator limits) is achievable, and our configuration can be modeled as a Dubin's car, which is controllable in the plane~\cite{LaValle2006PlanningAlgorithms}.}.
Appropriate choices of forces $f_i$, $i\in[1,N]$, for each individual module can generate desired surge (forward, or $y$ axis) forces and yaw torques on the configuration, which is sufficient to control it in the plane. No express control is given along the $x$ axis, so disturbances along the $x$ axis are assumed to be small. 

Define $v_y$ and $a_y$ as the velocity and acceleration of the COM along the $y$ axis, respectively, and $\Omega$ and $\alpha$ as the angular velocity and acceleration of the structure, respectively. We can then write the dynamics of the configuration as in~\eqref{eq:eom-y} and~\eqref{eq:eom-yaw}, where $x_i$ is the distance from the COM to module $i$ along the $\hat{x}_S$ axis, $C_L$ and $C_R$ are drag coefficients, and $m$ and $I$ are the mass and moment of inertia of the configuration, respectively. 

\begin{align}
    m a_y &= \sum_{i} f_{i} \hphantom{x_{i}} - C_L \sign{(v_y)} v_y^2  \label{eq:eom-y}\\
    I \alpha &= \sum_{i} f_{i}x_{i} - C_R \sign{(\Omega)} \Omega^2 \label{eq:eom-yaw}
\end{align}

Equations~\eqref{eq:eom-y} and~\eqref{eq:eom-yaw} can be rewritten in matrix form by defining the structural matrix $P$ as in~\eqref{eq:structuralMatrix}. This gives~\eqref{eq:eom-mat1}, where $\vec{f} = [\begin{matrix} f_1 & f_2 & \hdots & f_N \end{matrix}]^T$ is the vector of module forces. The right hand side of~\eqref{eq:eom-mat1} then represents the surge force and yaw torque that need to be applied to the structure to generate a desired surge velocity $v_c$ and to track a desired yaw angle $\Theta$. 

\begin{equation} \label{eq:structuralMatrix}
  P = \begin{bmatrix} 1 & 1 & \hdots & 1 & 1 \\ x_1 & x_2 & \hdots & x_{N-1} & x_N \end{bmatrix}
\end{equation}
\begin{equation} \label{eq:eom-mat1}
  P\vec{f} = \begin{bmatrix}  m a_y + C_L\abs{v_c}v_c \\ I\alpha + C_R\abs{\Omega}\Omega \end{bmatrix}
\end{equation}

For the purposes of control, we consider a case in which the desired surge velocity is (relatively) constant over time, while the desired yaw angle is varied to steer. This allows us to assume steady-state in the velocity equation and set $a_y = 0$. Eq.~\eqref{eq:eom-mat1} can then be solved for the individual module forces $\vec{f}$, as in~\eqref{eq:forces}, by using the Moore-Penrose pseudo-inverse $P^+ = P^T(PP^T)^{-1}$.

\begin{equation} \label{eq:forces}
    \vec{f} = P^+\begin{bmatrix}  C_L\abs{v_c}v_c \\ I\alpha + C_R\abs{\Omega}\Omega \end{bmatrix}
\end{equation}

Using~\eqref{eq:forces} to distribute forces among the modules results in a linear distribution along the $x$-axis of the configuration; a similar approach was used by Gabrich to distribute forces in a configuration of docked quadrotors~\cite{Gabrich2020ModQuad-DoF:Quadrotors}. Other distributions are possible, but a linear distribution most closely matches the internal dynamics that would be observed if the configuration were a single rigid body. As discussed in Sec.~\ref{sec:intro}, Modboat modules are docked using passive magnets that allow rotation, and significant intra-configuration forces can cause modules to undock. Maintaining a rigid-body force distribution minimizes such forces and oscillation between neighboring modules. 


\subsection{Control Input} \label{sec:control}

Eq.~\eqref{eq:forces} allows the configuration of Modboats to track a desired surge velocity $v_d$ and and yaw angle $\theta_d$. Surge motion is assumed to occur at steady-state on average, so it should be enough to use the desired velocity $v_d$ as the commanded velocity $v_c$ in~\eqref{eq:forces}. In practice, it is observed that this is not sufficient, however, so $v_d$ is instead used as a feedforward term, and a PD controller acts as an artificial acceleration to  adjust the commanded velocity $v_c$. The controller is given in~\eqref{eq:errVel} and~\eqref{eq:vCommand}, where $v_{obs}$ is the observed surge velocity of the configuration.
\begin{align}
    e_{v} &= v_{d} - v_{obs} \label{eq:errVel}\\ 
    v_c &= v_d + \int_{0}^t \left (K_{pv} e_{v} + K_{dv} \frac{de_{v}}{d\tau} \right ) d\tau \label{eq:vCommand}
\end{align}

The yaw angle can be commanded by a standard PD control loop on the angular acceleration, as given in~\eqref{eq:errYaw} and \eqref{eq:pidYaw}, where $\Theta$ and $\Theta_{des}$ are the observed and desired yaw, respectively. The angular velocity $\Omega$ used in~\eqref{eq:forces} is then the observed angular velocity of the configuration. 
\begin{align}
    e_{\Theta} &= \Theta_{des} - \Theta \label{eq:errYaw}\\ 
    \alpha &= K_{p\Theta} e_{\Theta} + K_{d\Theta} \frac{de_{\Theta}}{dt} \label{eq:pidYaw}
\end{align}


\subsection{Drag Coefficients} \label{sec:drag}

Using~\eqref{eq:forces} to drive the Modboat configuration requires knowledge of the constants relevant to the system, namely the moment of inertia $I$ and the linear and angular drag coefficients $C_L$ and $C_R$. The moment of inertia for an individual boat $I_i$ can be calculated from its Solidworks model files, and the total moment of inertia $I$ can then be calculated in a straightforward way via the parallel axis theorem for any configuration. 

\begin{figure}[t]
    \centering
    \subfloat[\label{fig:linearDrag}]{\includegraphics[width=0.95\linewidth, trim={0.5cm 0 0.5cm 0.25cm}, clip]{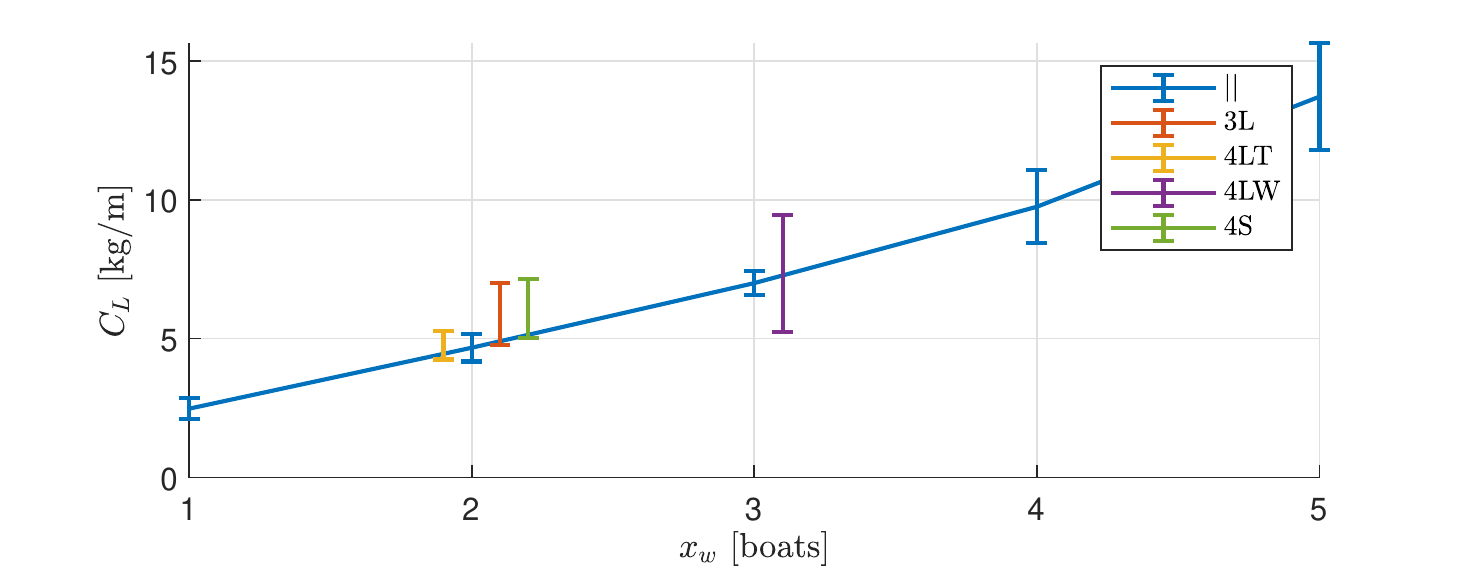}
       }
          \hfill
    \subfloat[\label{fig:angularDrag}]{\includegraphics[width=0.95\linewidth, trim={0.5cm 0 0.5cm 0.25cm}, clip]{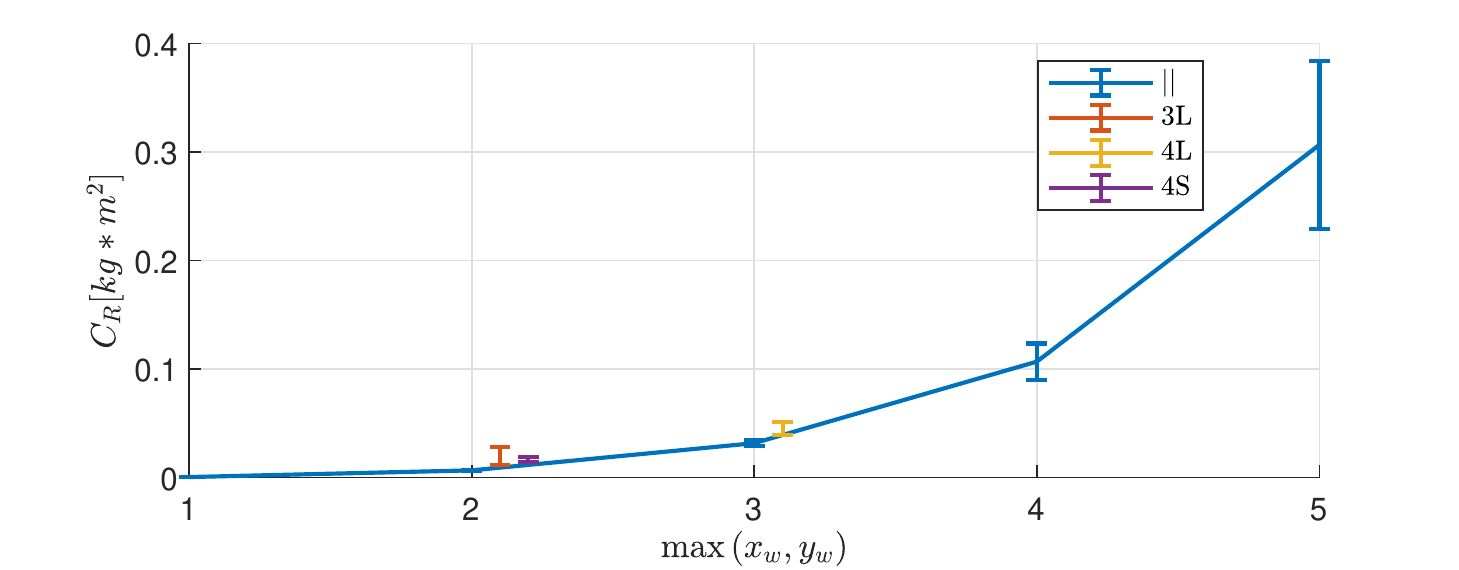}
        }
    \caption{Experimentally determined drag coefficients (a) $C_L$ and (b) $C_R$ vs. boat projection, as defined in~\eqref{eq:eom-y} and~\eqref{eq:eom-yaw}. Labels $x_w$ and $y_w$ indicate the width along the $x$ and $y$ axes, respectively. Connected plot shows results for parallel configurations, while the single bars indicate non parallel configurations shown in Fig.~\ref{fig:diagramsForDrag}, offset slightly along the $x$ axis for clarity. Error bars indicate one standard deviation.}
    \label{fig:dragCoeffs}
\end{figure}

The drag coefficients $C_L$ and $C_R$ can be experimentally calculated; a linear(angular) impulse is delivered to the configuration, and the resulting linear(angular) velocity is tracked. A nonlinear least-squares fit to a quadratic drag model then gives $C_L$ and $C_R$ when the moment of inertia and mass are known. The resulting drag coefficients are shown in Fig.~\ref{fig:dragCoeffs} and given in Table~\ref{tab:parameters} for \textit{parallel configurations}, where $C_L$ is shown to be roughly linear and $C_R$ is roughly quadratic. 

\begin{figure}[t]
    \centering
    \subfloat[\label{fig:3L}]{\includegraphics[page=4, width=0.2\linewidth]{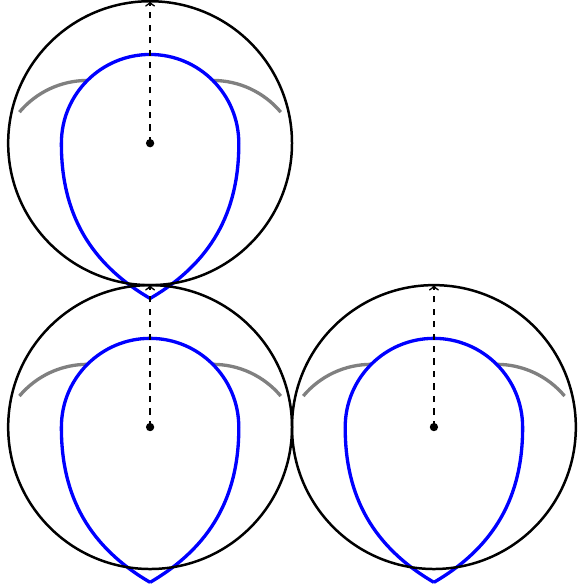}
      }
      \hfill
    \subfloat[\label{fig:4S}]{\includegraphics[page=7, width=0.2\linewidth]{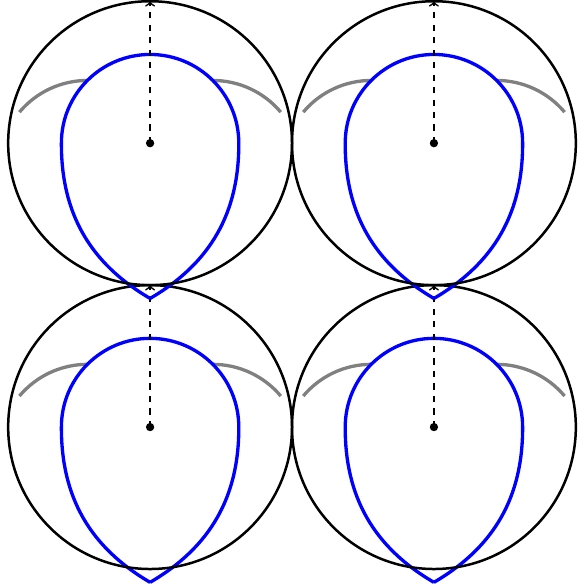}
      }    
      \hfill
    \subfloat[\label{fig:4LT}]{\includegraphics[page=5, width=0.2\linewidth]{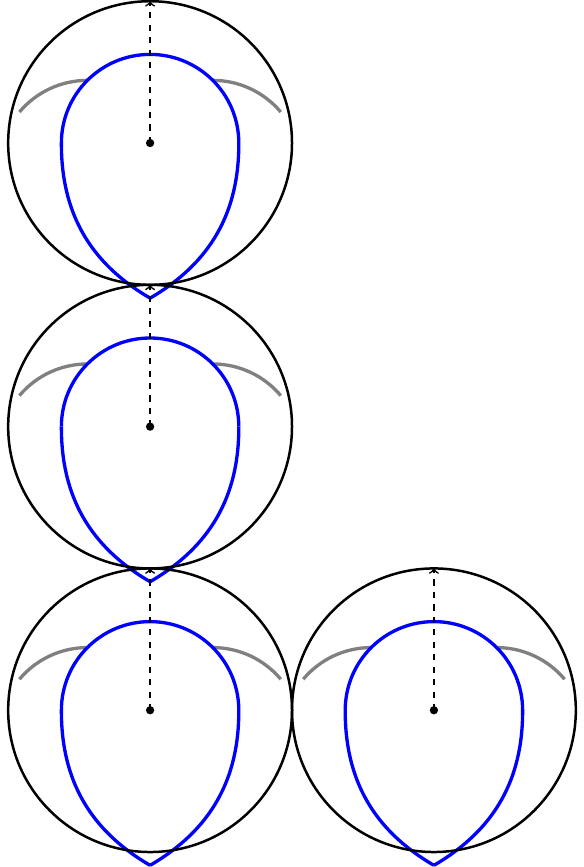}
      }
      \hfill
    \subfloat[\label{fig:4LW}]{\includegraphics[page=6, width=0.2\linewidth]{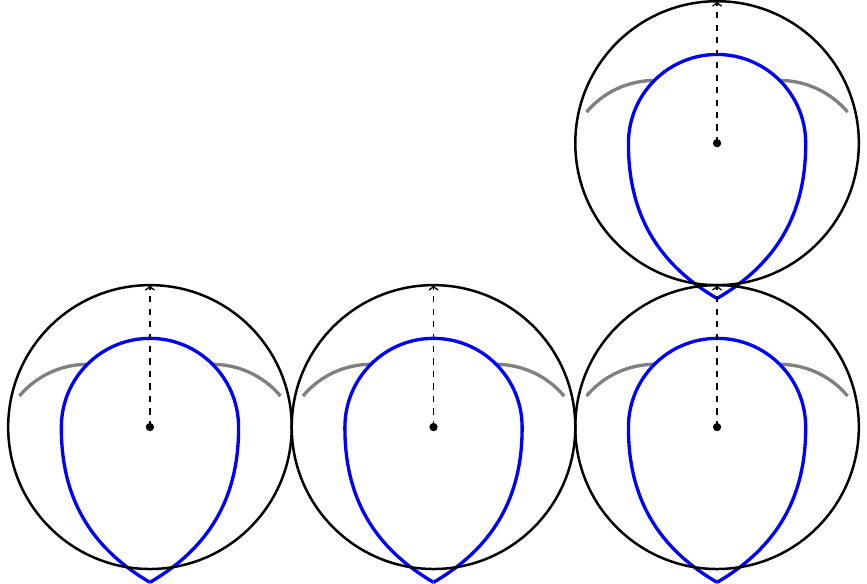}
      }
    \caption{Diagrams of Modboat configurations evaluated for drag coefficients and shown in Fig.~\ref{fig:dragCoeffs}. (a) $3L$, (b) $4S$, (c) $4LT$, (d) $4LW$. Note that while $4LT$ and $4LW$ are distinct for linear motion, as in Fig.~\ref{fig:linearDrag}, they are identical for rotation, as in Fig.~\ref{fig:angularDrag}.}
    \label{fig:diagramsForDrag}
\end{figure}

However, it is clearly impractical to use this method for arbitrary configurations, since it requires experimental evaluation for \textit{every} possible configuration. Two factors combine to allow us to avoid such labor intensive experimentation. First, experimental evaluation on a small selection of non-parallel configurations (see Fig.~\ref{fig:diagramsForDrag}) shows that their drag coefficients correspond reasonably well to the value given based on their \textit{projection} for parallel configurations, as shown in Fig.~\ref{fig:dragCoeffs}. Second, as we will show in Sec.~\ref{sec:experiments}, even using significantly mismatched drag coefficients results in only minor performance penalties after feedback is applied.

Thus, we project the configuration onto its axes and use the drag coefficients predicted for parallel configurations when applying~\eqref{eq:forces} for control. For linear drag $C_L$, this is the width along the $\hat{x}_S$ axis $x_w$. For rotational drag $C_R$ we use the maximum width along either the $\hat{x}_S$ or the $\hat{y}_S$ axes.

\begin{table}[t]
    \centering
    \caption{Mass and inertia for each parallel boat configuration.}
    \begin{tabular}{rl|c|c|c|c|c}
    \toprule
    $N$ & Boats & 1 & 2 & 3 & 4 & 5 \\ \midrule
    $M$ & $[\si{kg}]$   &  0.66 & 1.32 & 1.98 & 2.64 & 3.30\\
    $I$ & $[\si{g*m^2}]$ &  2.05 & 11.8 & 36.8 & 84.8 & 164 \\
    $C_L$ & $[\si{kg/m}]$ & 2.48 & 4.67 & 7.00 & 9.75 & 13.7 \\
    $C_R$ & $[\si{g*m^2}]$ & 0.40 & 6.50 & 32.0 & 107 & 307 \\ \bottomrule
    \end{tabular}
    \label{tab:parameters}
\end{table}


\section{Waveform} \label{sec:waveform}

In Sec.~\ref{sec:dynamics}, we posited a set of modules that can produce positive and negative thrust along their $y$-axes. It remains to be shown how this can be achieved with the Modboat, and how to guarantee that the resulting system is collision-free under the control implementation of Sec.~\ref{sec:dynamics}. 

When Modboat modules swim alone, their top body section acts as a inertial rotor to allow the propulsive bottom body to rotate in the water~\cite{modboatsOnline}. Nevertheless, some of the motion goes into the top body, and this must be accounted for in single boat control schemes~\cite{Knizhnik2020a,Knizhnik2021a}. However, when multiple Modboats are docked together the dock acts to significantly reduce the rotation of the top body. This allows us to equate the orientation of the bottom body $\theta$ and the motor angle $\phi$ (shown in Fig.~\ref{fig:diagramNP}), and consider the input waveform directly as a measure of propulsion.

Consider a set of Modboats executing the waveform given in~\eqref{eq:waveformNew}, where the centerline is given by $\phi_0$ and the amplitude by $A$, and the subscript $i$ indicates the particular module. Wherever necessary, we assume that a waveform is executed with its set of parameters for a single cycle, and then the parameters are updated for the next cycle; a subscript indicating the cycle is omitted for clarity, however. Note that the angular frequency $\omega$ is constant for all modules for concurrency of decision\footnote{In practice $\omega$ is also held constant between cycles, but it need not be.}.
\begin{equation}\label{eq:waveformNew}
        \phi_i(t) = (\phi_0)_i + A_{i} \cos{(\omega t)}\cos{\left ((\phi_0)_{i} \right )}
\end{equation}

Over a complete cycle (of length $T$, where $T$ is the period corresponding to angular frequency $\omega$) we note that~\eqref{eq:waveformNew} is symmetric about $\phi_0$. As has been shown in prior work~\cite{Knizhnik2020a}, under a symmetric waveform lateral forces cancel and the Modboat produces an average force along the direction given by $\phi_0$. So if $\phi_0 = 0~\si{rad}$ we produce positive force along the $y$ axis, and can vary its magnitude by varying $A$.

To achieve negative forces, we allow $\phi_0 = \pi~\si{rad}$; this reverses the direction of the centerline and produces negative thrust along the same axis. The $\cos{\left ((\phi_0)_{i} \right )}$ term in~\eqref{eq:waveformNew} has the effect of reversing the sign of the amplitude when $\phi_0 = \pi~\si{rad}$, which minimizes the discontinuity that occurs when transitioning from $\phi_0 = 0~\si{rad}$ to $\phi_0 = \pi~\si{rad}$. We cannot completely remove the discontinuity, however, and its effects will be considered in Sec.~\ref{sec:experiments}.

\begin{figure}[t]
    \centering
    \includegraphics[width=\linewidth]{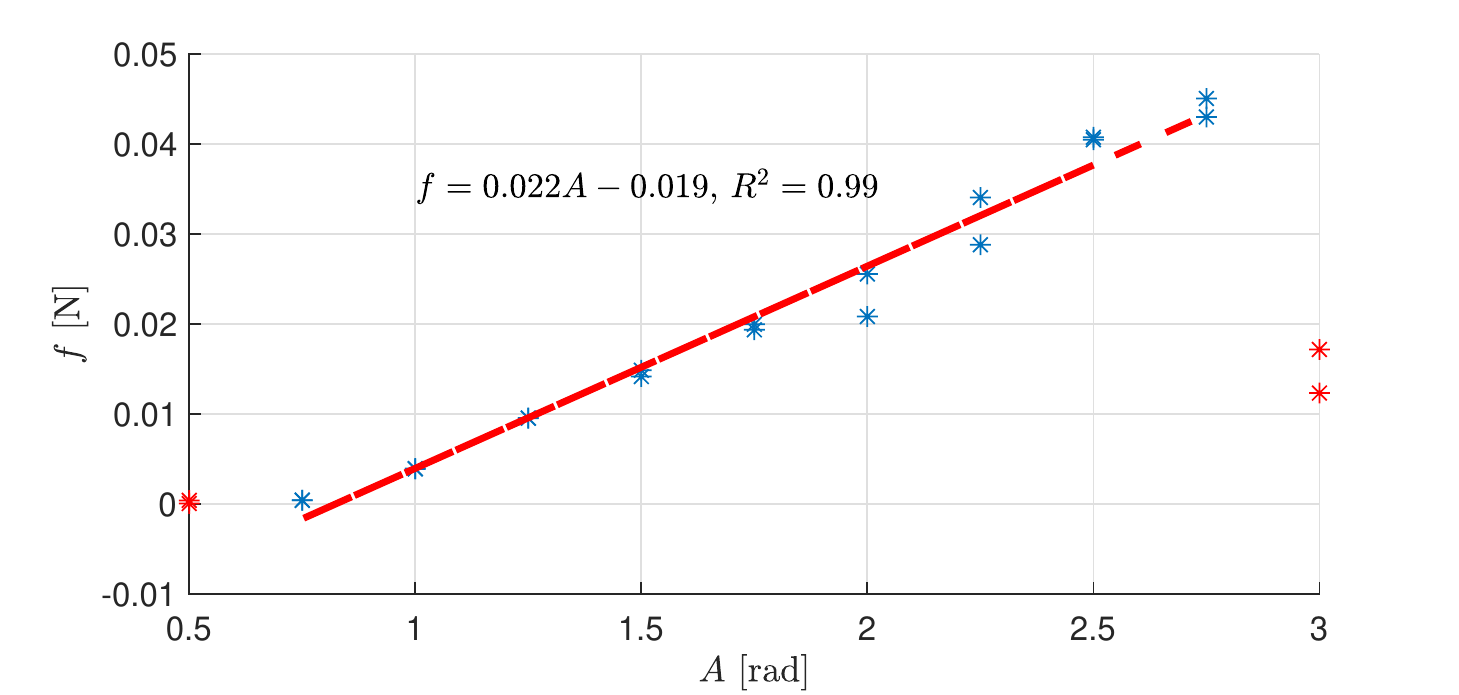}
    \caption{The experimentally determined force vs. amplitude curve for a period of oscillation $T=1.5~\si{s}$. Data points are shown in blue, and red stars indicate poor force generation due to incomplete flipper activation (at low amplitudes) or significant reverse thrust (at high amplitudes).}
    \label{fig:forceFromAmp}
\end{figure}

We can experimentally determine the thrust produced using this waveform. Three Modboats were set up in a parallel configuration, and the center module was set to execute~\eqref{eq:waveformNew} with $\phi_0 = 0$ and varying amplitudes. The steady-state velocity was then used in combination with a measured drag coefficient to compute the mapping given in Fig.~\ref{fig:forceFromAmp}.

Fig.~\ref{fig:forceFromAmp} shows the resulting mapping $f = f_0(A)$ for $T=1.5~\si{s}$, which is linear within the range $A\in[0.75,2.75]~\si{rad}$. Below $0.75~\si{rad}$ the flippers do not fully open, so negligible thrust is produced. Above $2.75~\si{rad}$, the tail rotates enough to produce significant reverse thrust during a portion of the cycle. Thus we intentionally limit the maximum allowable amplitude to $2.5~\si{rad}$\footnote{Although thrust is maintained up to $2.75~\si{rad}$, in practice amplitudes higher than $2.5~\si{rad}$ cause the configuration to shake internally.}. 

Thus, a docked Modboat executing the waveform given in~\eqref{eq:waveformNew} for $\phi_0 \in \{0,\pi\}~\si{rad}$ acts as a module that can produce varying thrust along its $y$ axis when averaged over a period of length $T$, with the amplitude of the waveform $A$ as the input variable. 


\section{Hydrodynamic Interactions} \label{sec:hydrodynamics}

As noted in Sec.~\ref{sec:dynamics}, it is difficult to accurately model hydrodynamic interactions between swimming bodies located close to one another. Significant work has been done in the literature to determine the effects of swimmers in a configuration~\cite{Weihs1973HydromechanicsSchooling,Ashraf2017SimpleSchooling,Maertens2017OptimalSwimmers,Ramananarivo2016FlowFlight,Khalid2018OnConfiguration, Becker2015HydrodynamicSwimmers, Liao2003FishActivity,Li2017NumericalModel,Li2020VortexFish,Li2021UsingFish}, but --- to the best of the authors' knowledge --- none has achieved a model simple enough for effective control use. This is especially difficult for flapping swimmers, where the time scale of the wake behavior is comparable to that of the actuation. 

In our prior work considering only parallel configurations~\cite{Knizhnik2022AmplitudeModboats} we assumed no hydrodynamic interactions between horizontal neighbors for ease of control, and the resulting performance was reasonable enough to justify this simplifying assumption. For non-parallel configurations, however, this assumption can no longer be justified, as performance decreases significantly when it is made\footnote{E.g. yaw tracking error while swimming for the $L$ configuration increases by $89\si{\%}$, with mean $0.21~\si{rad}$ and IQR $[0.15,0.26]~\si{rad}$. Compare with mean $0.11~\si{rad}$ and IQR $[0.051,0.17]~\si{rad}$ in Table~\ref{tab:resultsRMS} when hydrodynamic effects are modeled.}. Some hydrodynamic modeling is therefore necessary, but it is desirable that the model be readily integrated into the dynamics and control method presented in Sec.~\ref{sec:dynamics}, which has already been validated~\cite{Knizhnik2022AmplitudeModboats}.

The most obvious wake interaction to consider when extending to arbitrary configurations is the interaction between vertical neighbors. Since the Modboat's thrust is based on the relative velocity between its flippers and the fluid, swimming in the rearward wake of another boat would be expected to decrease the thrust produced. For a first approximation we consider interactions only along the $y$ axis of Fig.~\ref{fig:diagramNP} (i.e. the wake does not spread laterally). This is reasonable since --- in any case --- we are restricting thrust to be along the $y$ axis.

We consider the simplest model of hydrodynamic interactions: a linear thrust loss due to interactions with the wake of neighbors in front. Thus, if Sec.~\ref{sec:waveform} introduced the thrust map $f = f_0 (A)$ for a Modboat swimming alone, then under the thrust loss model each boat $i$ would produce $f_i = \alpha_i f_0 (A_i)$ for some constant $\alpha \leq 1$. 

\begin{figure}[t]
    \centering
    \includegraphics[width=\linewidth]{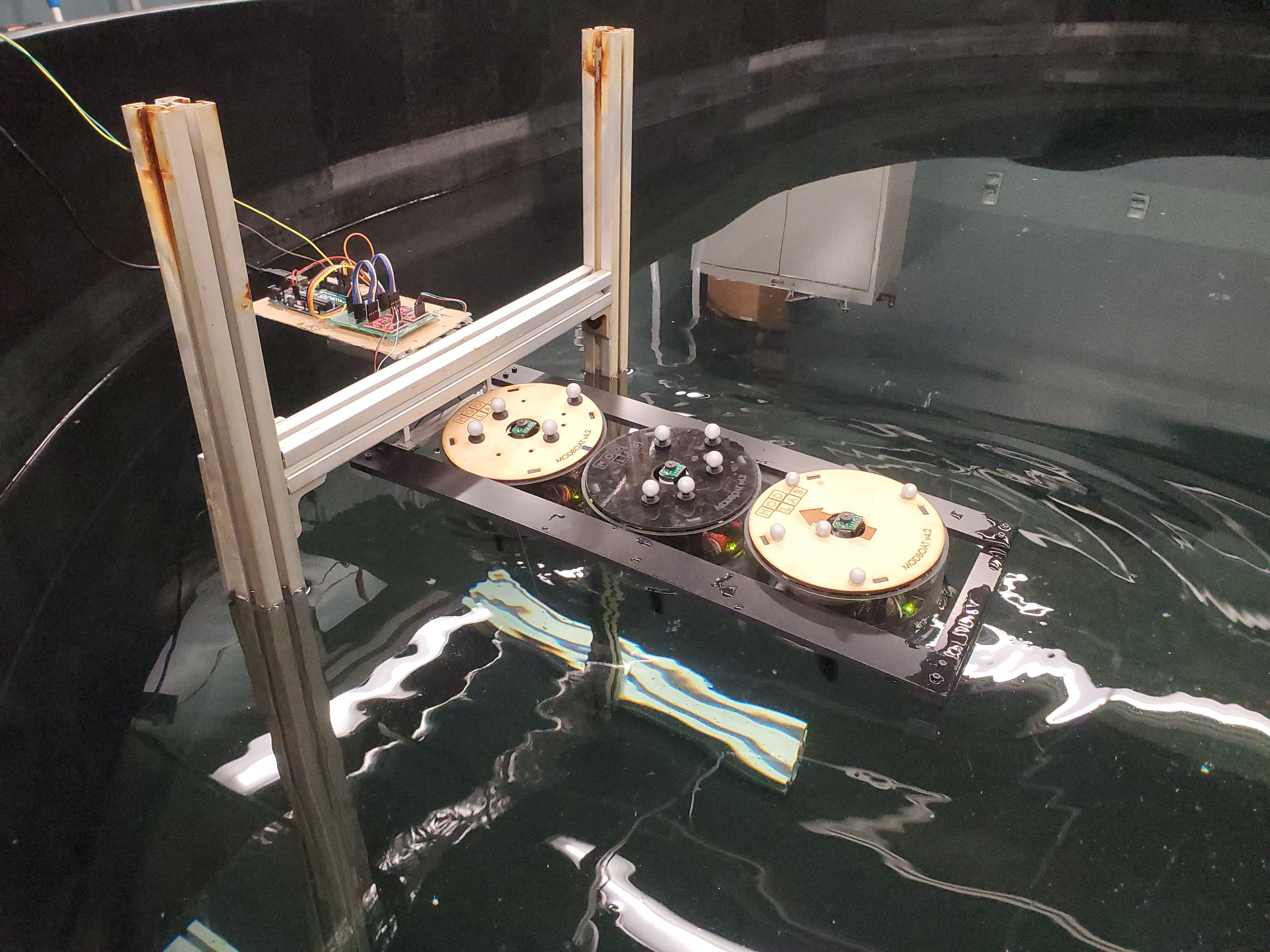}
    \caption{A photo of three Modboats in the thrust stand configuration. One, two, and then three Modboats were actuated with identical amplitude $A$ in order to evaluate the contribution of each to the overall thrust of the group.}
    \label{fig:thrustStand}
\end{figure}

\begin{figure}[t]
    \centering
    \includegraphics[width=\linewidth]{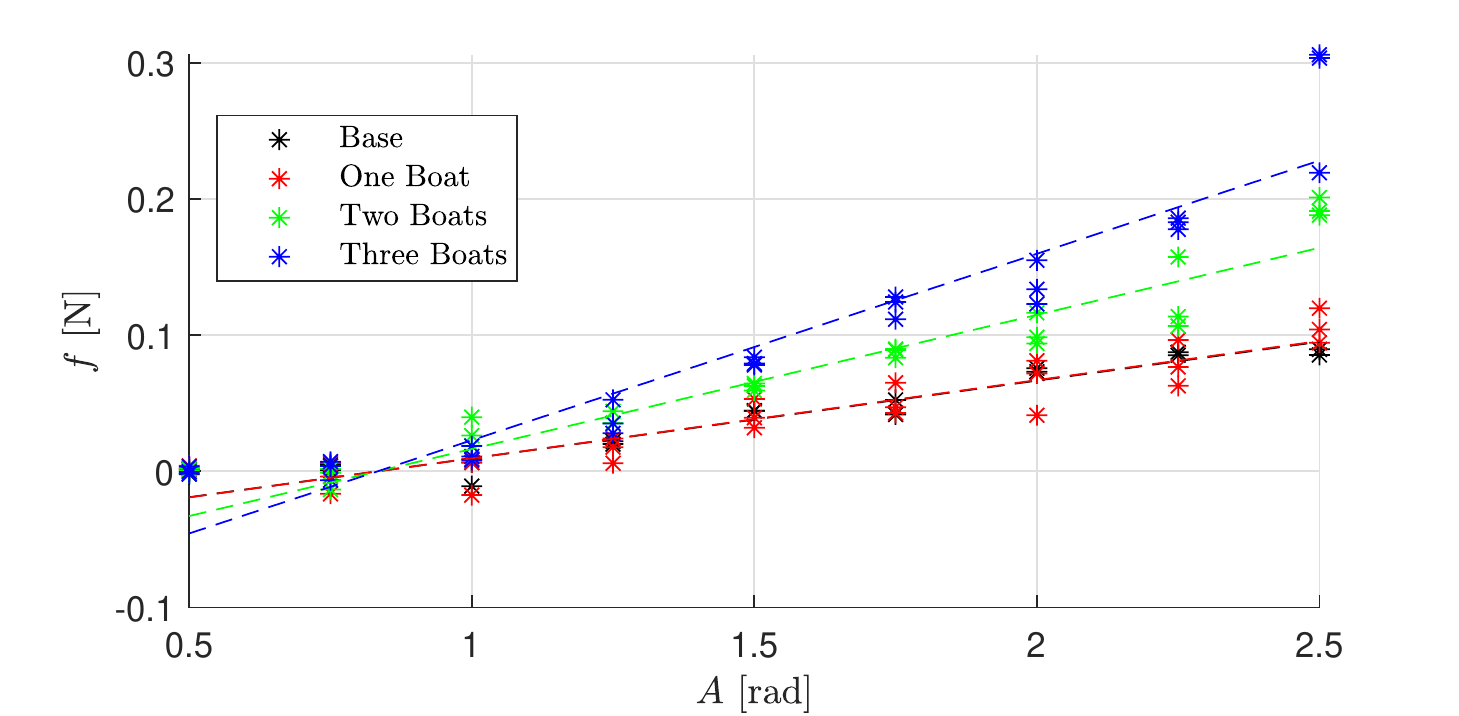}
    \caption{Average thrust produced as a function of oscillation amplitude $A$ for one, two, and three docked boats as in Fig.~\ref{fig:thrustStand}. A baseline is added in which one Modboat was actuated without the other two attached. The lines of best fit are matched to have the same $x$-intercept as the baseline.}
    \label{fig:thrustLoss}
\end{figure}

Since we are considering wakes that move only vertically, it is sufficient to consider a column of the configuration to determine the coefficients $\alpha_i$. We can therefore experimentally model this thrust loss by placing a column of Modboats on a thrust stand and measuring the thrust produced at various oscillation amplitudes. All the Modboats were given the same amplitude $A$; first one boat only was actuated, then the first two, and then three\footnote{More than three boats can be tested this way, but we did not test further due to practical limitations.}. Fig.~\ref{fig:thrustStand} shows the experimental setup, and the results are given in Fig.~\ref{fig:thrustLoss}.

The data shown in Fig.~\ref{fig:thrustLoss} is reasonably linear, so we are justified in pursuing a linear thrust model. To fit the coefficients $\alpha$, we found the line of best fit for the \textit{base} case (with slope $m_0$) in Fig.~\ref{fig:thrustLoss}, and then determined the best fit slopes for lines sharing an $x$ intercept for the remaining data sets\footnote{We note that the best fit line for the thrust of the base case is approximately 3x bigger than the equivalent line from Fig.~\ref{fig:forceFromAmp}. This may be attributed to the low-cost thrust stand used in Fig.~\ref{fig:thrustStand} and/or to poor filtering of the data. Nevertheless, only the relative slopes of the lines in Fig.~\ref{fig:thrustLoss} are used, with the absolute magnitude taken from Fig.~\ref{fig:forceFromAmp}}. Then the $\alpha$ coefficients can be found via~\eqref{eq:findAlpha}, where $\alpha(k)$ indicates the coefficient $\alpha$ for a boat that has $k-1$ Modboats in front of it. The results are given in Table~\ref{tab:thrustProps}; wake effects reduce the rearward Modboats' thrusts by around $30\si{\%}$.

\begin{equation} \label{eq:findAlpha}
    \alpha(k) = \frac{m_k - m_{k-1}}{m_{0}} \quad k \in [1,3]
\end{equation}

\begin{table}[t]
    \centering
    \caption{Thrust proportions $\alpha$ and wake compensation coefficients $\gamma$ for boats swimming in the rear.}
    \begin{tabular}{c|c|c|c} \toprule
        Boat    & One & Two & Three \\ \midrule
        $\alpha$ & 1.0 & 0.72 & 0.67 \\
        $\gamma$ & 1.0 & 1.49 & 1.07 \\ \bottomrule
    \end{tabular}
    \label{tab:thrustProps}
\end{table}

Since the thrust map $f_0(A)$ is linear, under this model we can compensate for thrust loss by increasing the commanded amplitude $A_i$ by a factor $1/\alpha_i$. It is therefore convenient to define a wake compensation factor using the coefficient $\alpha$. Recall that, under the wake model we have assumed, the thrust reduction for any boat $i$ can depend only on $k$, where there are $k-1$ boats in front of it. 

\begin{definition}[Wake compensation Factor] \label{def:wakeCompFactor}
Let the \textit{wake compensation factor} $\gamma$ of a boat $i$ be given by:

\begin{equation} \label{eq:gamma}
    \gamma(k) = \frac{\alpha(k-1)}{\alpha(k)} = \frac{A(k)}{A(k-1)} \quad k \in \mathbb{Z}^*
\end{equation}

Then it is always true that for a boat $(i,j)$  that has a rearward value $k$, $A(i,j) = \gamma(k)A(i,j+1)$. That is $\gamma$ defines the ratio between the amplitude of a boat and its forward neighbor. The wake compensation factors measured for the Modboats are given in Table~\ref{tab:thrustProps}
\end{definition}

While the measurements and models presented in this section cannot capture the entirety of the hydrodynamic interactions and are a significant oversimplification, the results presented in Sec.~\ref{sec:experiments} demonstrate that even this first order approximation is sufficient --- in conjunction with feedback control --- to provide a reasonable model of system behavior.


\section{Avoiding Unintentional Undocking} \label{sec:selfCollisionNP}

In Sec.~\ref{sec:dynamics} we presented a controller for an arbitrary configuration of modules that can produce thrust along a single aligned axis, and in Sec.~\ref{sec:waveform} we showed that Modboats can act like such a module when docked and averaged over a full cycle of length $T$. 

The challenge for control of a configuration of Modboats, however, lies in the multiplexed function of the bottom body tail. As shown in Fig.~\ref{fig:diagramNP}, the bottom body of the Modboat (blue) and flippers (gray) are fully contained within the footprint of the top body (black) except for the tip of the tail. This ensures that the flippers of neighboring modules cannot mechanically interact, but the tails can be used to undock from neighboring modules by bringing them into contact (see Fig.~\ref{fig:diagramNP})~\cite{Knizhnik2021}, which is essentially self-collision within the configuration. This is advantageous because docking and undocking can be performed without additional actuation, but introduces a complex constraint when swimming as a unit. 

It therefore remains to show that the waveform defined in Sec.~\ref{sec:waveform} is \textit{sufficient} to avoid unintentional self-collisions for all allowable inputs. More general solutions likely exist but are complex to find and define; they are deferred to future work.

Formally, we can define some consequences of the application of the waveform~\eqref{eq:waveformNew} under the restriction $\phi_0 \in \{0,\pi \}~\si{rad}$. Assumption~\ref{asp:phaseLock} stems directly from the lack of a phase offset term in~\eqref{eq:waveformNew}. Similarly, assumption~\ref{asp:forward} formalizes the choice of $\phi_0 \in \{0, \pi \}~\si{rad}$. 

\begin{assumption}[Phase Lock] \label{asp:phaseLock}
All boats are \textbf{in phase} with one another for all time, i.e there is no phase offset in~\eqref{eq:waveformNew} and $\omega$ is the same for all boats. Control decisions are made concurrently for all boats in the configuration at the end of each cycle.
\end{assumption}

\begin{assumption}[Forward/Reverse] \label{asp:forward}
All boats can choose the centerline of rotation $(\phi_0)_i$ to be either $0~\si{rad}$ (forward) or $\pi~\si{rad}$ (reverse) in each cycle. No other angles are allowed.
\end{assumption}

We can then propose Theorem~\ref{th:selfCollision} in conjunction with the rearward wake compensation factor in Definition~\ref{def:wakeCompFactor}.

\begin{theorem}[No Unintentional Undocking]\label{th:selfCollision}
When swimming with waveform given by~\eqref{eq:waveformNew} under assumptions~\ref{asp:phaseLock} and~\ref{asp:forward}, we can guarantee that no unintentional undocking events will occur within an \textit{arbitrary} configuration of docked Modboats as long as the maximum wake compensation factor $\gamma$ is sufficiently small.
\end{theorem}

\begin{figure}[t]
    \centering
    \includegraphics[page=3, width=\linewidth]{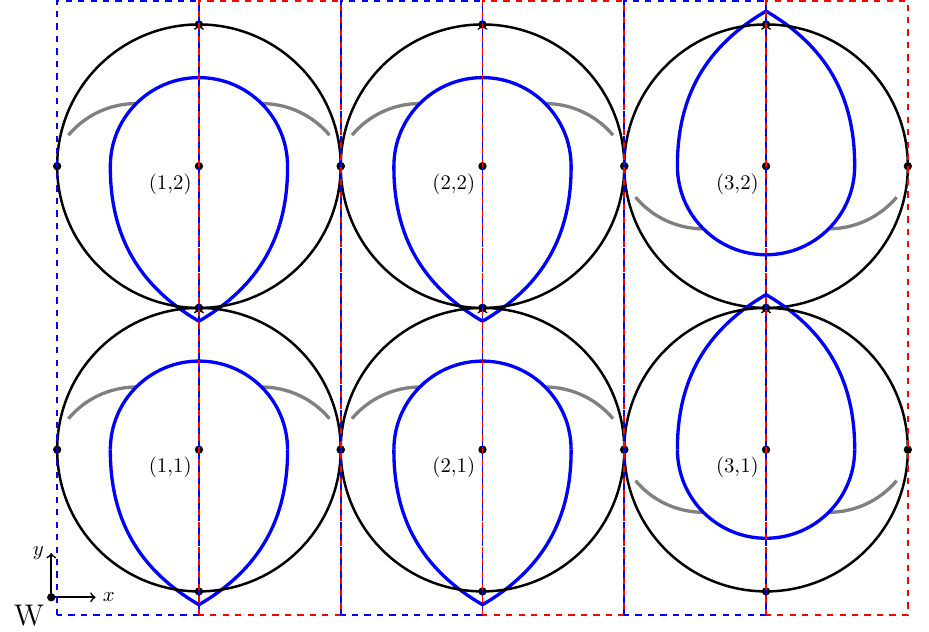}
    \caption{configuration of six docked Modboats, with red-blue tiling used in proof for Theorem~\ref{th:selfCollision}. Modboats $(3,1)$ and $(3,2)$ are shown in the reverse paddling configuration ($\phi = \phi_0 = \pi$ in~\eqref{eq:waveformNew}), while the rest are in the forward configuration ($\phi = \phi_0=0$), shown at $t = 0$.}
    \label{fig:diagramForProofNP}
\end{figure}

\begin{proof}
Construct a configuration of Modboats as in Fig.~\ref{fig:diagramForProofNP}. The tail protrusion is small enough to guarantee no collisions with diagonal neighbors, so we must consider only collisions with horizontal and vertical neighbors.

For \textbf{horizontal neighbors}, construct a lattice of neighboring red and blue regions, as shown in Fig.~\ref{fig:diagramForProofNP}. For any configuration width (along the $x$ axis), it is trivial to periodically tile. Consider any \textit{horizontal} row of the configuration at some time $t_0$ when $\cos{(\omega t_0)}=1$, and let $\phi_0 = 0$ for all boats. Then $\phi_i = (\phi_0)_i + A\cos{\left((\phi_0)_{i} \right )}$ $\forall i$, and 

\begin{equation*}
\phi_i = \begin{cases}
(\phi_0)_i & t = t_0 + T/4 \\
(\phi_0)_i - A\cos{\left((\phi_0)_{i} \right )} & t = t_0 + T/2 \\
(\phi_0)_i & t = t_0 + 3T/4 \\
(\phi_0)_i + A\cos{\left((\phi_0)_{i} \right )} & t = t_0 + T
\end{cases} \quad \forall i
\end{equation*}

Thus the tail segments are \textit{all} in red regions\footnote{The tail tip will eventually enter the neighboring boat's blue region. For the purposes of the proof, we consider a slightly interior point that remains within the red region.} when $t \in (t_0, t_0 + T/4) \cap (t_0 + 3T/4,t_0 + T)$, and blue regions when $t \in (t_0 + T/4, t_0 + 3T/4)$. Thus, at all times all tails occupy identically colored regions. By construction, no neighboring regions share a color, so no \textit{horizontal} collisions are possible.

\textbf{Vertical collisions} can be similarly considered. The linear distribution of forces created by the Moore-Penrose pseudo-inverse is \textit{guaranteed} to generate identical forces $f_i$ for boats with identical displacements $x_i$ (see Appendix~\ref{app:sameForces}). This guarantees that all boats in a vertical column will make the same choice of $\phi_0$. Collisions caused by boat $(i,j)$ selecting $\phi_0 = \pi~\si{rad}$ and its forward neighbor $(i, j+1)$ selecting $\phi_0 = 0~\si{rad}$ are thus impossible.

Additional consideration must be given to the wake compensation coefficient $\gamma$. For any boat $(i,j)$, we have $\phi_{(i,j)} = A$ and $\phi_{(i,j-1)} = \gamma A$ at $t = t_0$. If $\gamma$ is sufficiently large, we could have $\phi_{(i,j-1)} \approx \pi$ while $\phi_{(i,j)} \approx 0$, leading to a collision, and this occurs for $\gamma \geq 1.9$ (see Appendix~\ref{app:verticalNeighbors}). Table~\ref{tab:thrustProps} shows the maximum $\gamma$ in the configurations to be $1.49$, so collisions are impossible.
\end{proof}


\section{Experiments} \label{sec:experiments}

\begin{figure}[t]
    \centering
    \subfloat[\label{fig:4S1}]{\includegraphics[page=7, width=0.2\linewidth]{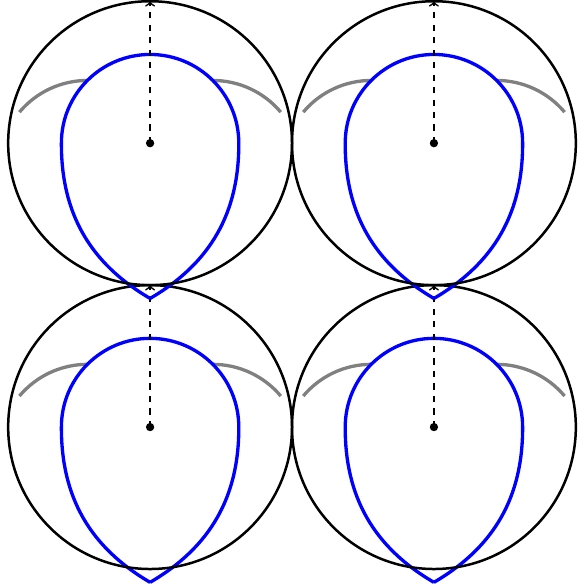}
      }
      \hfill
    \subfloat[\label{fig:4T}]{\includegraphics[page=8, width=0.3\linewidth]{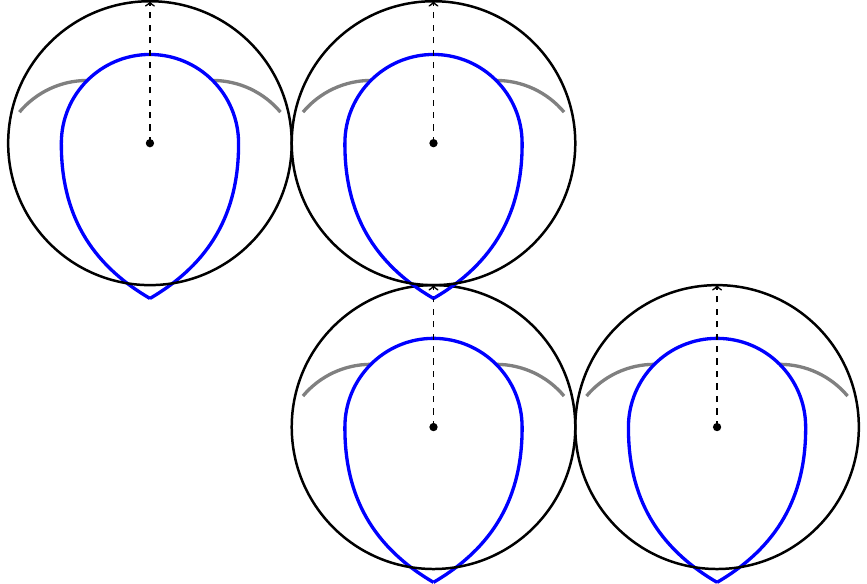}
      }    
      \hfill
    \subfloat[\label{fig:5P}]{\includegraphics[page=9, width=0.3\linewidth]{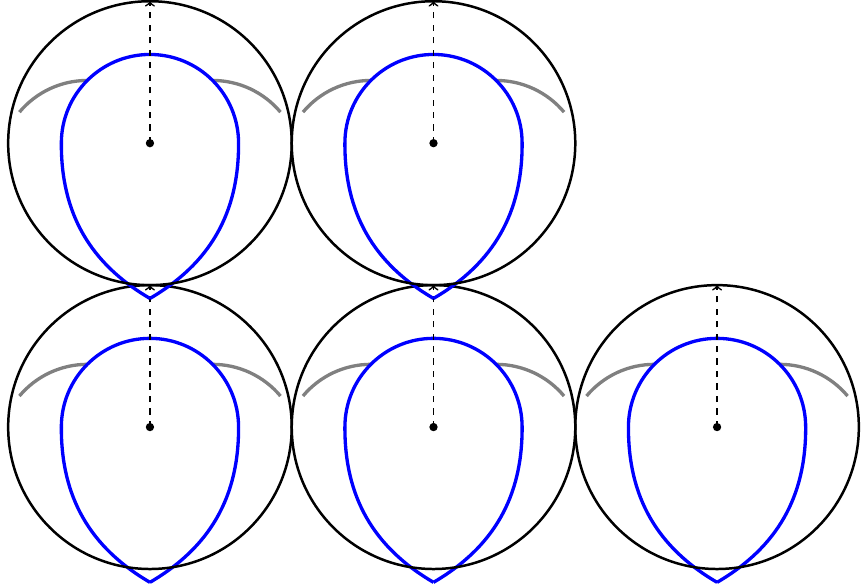}
      }
      \hfill
    \subfloat[\label{fig:5U}]{\includegraphics[page=10, width=0.3\linewidth]{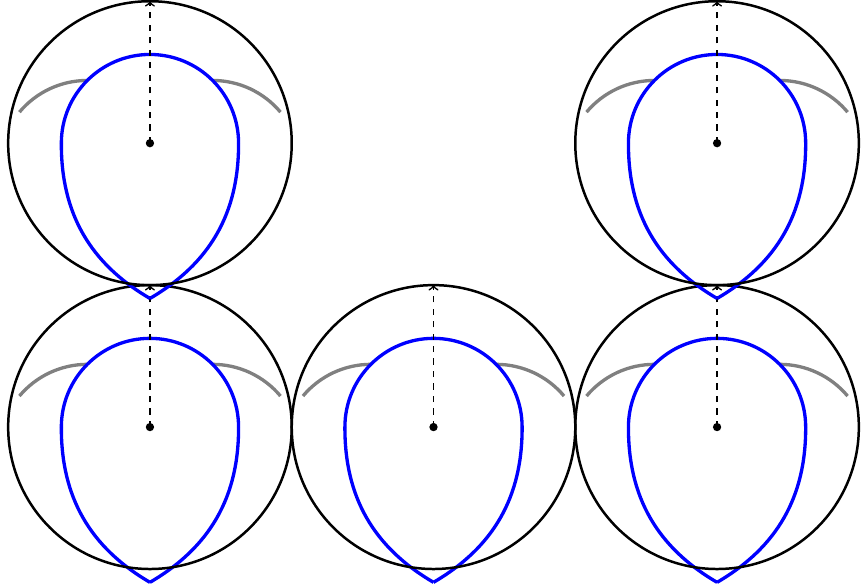}
      }
      \hfill
    \subfloat[\label{fig:5L}]{\includegraphics[page=11, width=0.3\linewidth]{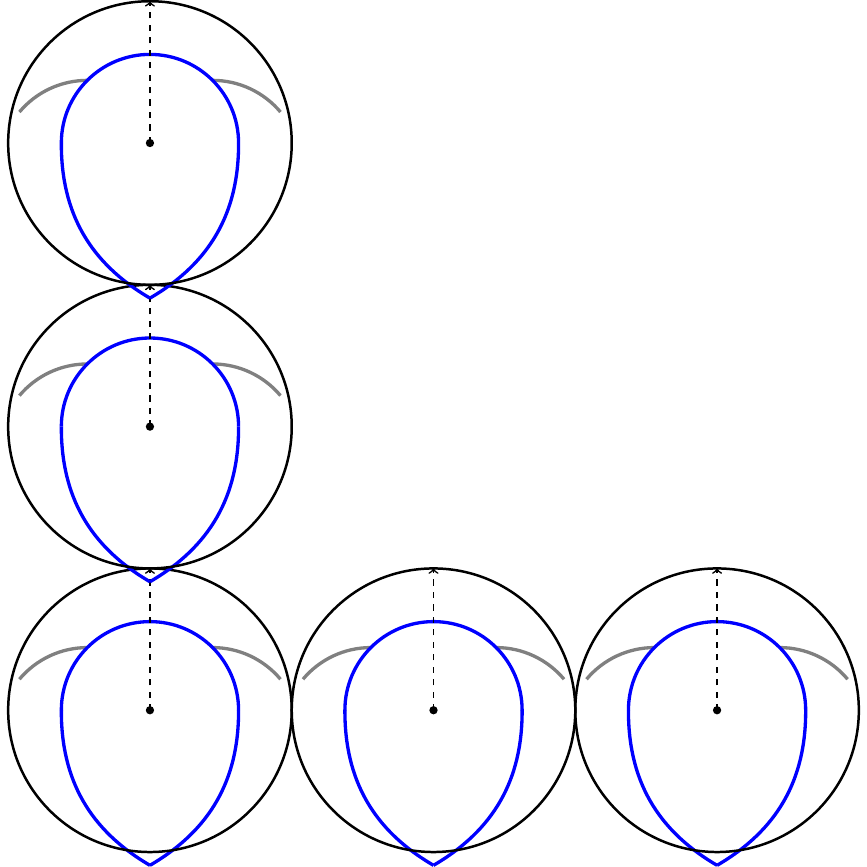}
      }
    \caption{Diagrams of non-parallel Modboat configurations used in experimental evaluation. We considered four boat (a) square and (b) tetromino shapes, and five boat (c) P, (d) U, (e) L configurations.}
    \label{fig:diagramsForExperiments}
\end{figure}

\begin{table}[t]
    \centering
    \caption{Controller coefficients used in \crefrange{eq:errVel}{eq:pidYaw}}
    \begin{tabular}{c|c|c|c} \toprule
       $K_{pv}~[\si{s^{-1}}]$ & $K_{dv}$ & $K_{p\Theta}~[\si{s^{-2}}]$  & $K_{d\Theta}~[\si{s^{-1}}]$  \\ \midrule
       $0.6$ & $0.1$ & $1.0$  & $0.2$  \\ \bottomrule
    \end{tabular}
    \label{tab:controller}
\end{table}

Modboat configurations were experimentally evaluated in a $4.5~\si{m} \times 3.0~\si{m} \times 1.2 ~\si{m}$ tank of still water, equipped with an OptiTrack motion capture system providing planar position, orientation, and velocity data at $120~\si{Hz}$. A MATLAB script calculated forces via~\eqref{eq:forces} and used the mapping in Fig.~\ref{fig:forceFromAmp} and wake-compensation parameters to determine the required amplitude and centerline for each boat. These parameters were communicated to each Modboat over WiFi, and an onboard ESP32-based controller executed the waveform in~\eqref{eq:waveformNew}.

\begin{table}[t]
    \centering
    \caption{Controller RMS performance as IQR. 
    Parallel configurations are summarized as \textbf{Par}. Non-parallel configurations are summarized as \textbf{NP}.}
    \setlength{\tabcolsep}{5.0pt}
    \begin{tabular}{c|c|c|c|c}
    
    \toprule
        ~ & \multicolumn{2}{c|}{ $v_y$ RMS $[\si{cm/s}]$} & \multicolumn{2}{c}{Yaw RMS $[\si{rad}]$}  \\
        ~ & Vel. only & Vel. \& Yaw & Yaw only & Vel. \& Yaw \\ \midrule
        2 & $[0.12,1.29]$ & $[0.25,0.89]$ & $[0.28\hphantom{0},0.36]$ & $[0.099, 0.21\hphantom{0}]$ \\
        3 & $[0.14,0.41]$ & $[0.27,0.48]$ & $[0.14\hphantom{0},0.17]$ & $[0.058, 0.17\hphantom{0}]$\\
        4 & $[0.15,0.29]$ & $[0.20,0.32]$ & $[0.10\hphantom{0},0.19]$ & $[0.035,0.077]$\\
        5 & $[0.19,0.34]$ & $[0.20,0.42]$ & $[0.073,0.10]$ & $[0.064,0.13\hphantom{0}]$\\ \midrule
        \textbf{Par.} & $[0.13, 0.38]$ & $[0.23, 0.42]$ & $[0.10\hphantom{0}, 0.24]$ & $[0.065, 0.15\hphantom{0}]$ \\ 
        \textbf{NP} & $[0.25, 0.39]$ & $[0.24, 0.63]$ & $[0.17\hphantom{0}, 0.23]$ & $[0.062, 0.16\hphantom{0}]$ \\ \midrule
        Square & $[0.21, 0.39]$ & $[0.41, 0.77]$ & $[0.22\hphantom{0}, 0.25]$ & $[0.074, 0.21\hphantom{0}]$ \\
        Tetr. & $[0.19, 0.28]$ & $[0.31, 0.41]$ & $[0.23\hphantom{0}, 0.28]$ & $[0.12\hphantom{0},  0.18\hphantom{0}]$ \\ 
        P & $[0.30, 0.42]$ & $[0.23, 0.37]$ & $[0.17\hphantom{0}, 0.21]$ & $[0.051, 0.12\hphantom{0}]$ \\ 
        U & $[0.30, 0.39]$ & $[0.23, 0.78]$ & $[0.18\hphantom{0}, 0.21]$ & $[0.035, 0.085]$ \\
        L & $[0.26, 0.40]$ & $[0.23, 0.51]$ & $[0.16\hphantom{0}, 0.17]$ & $[0.051, 0.17\hphantom{0}]$ \\ \bottomrule
    \end{tabular}
    \label{tab:resultsRMS}
\end{table}

\begin{figure}[t]
    \centering
    \includegraphics[width = \linewidth, trim={0 0 0 0.5cm}]{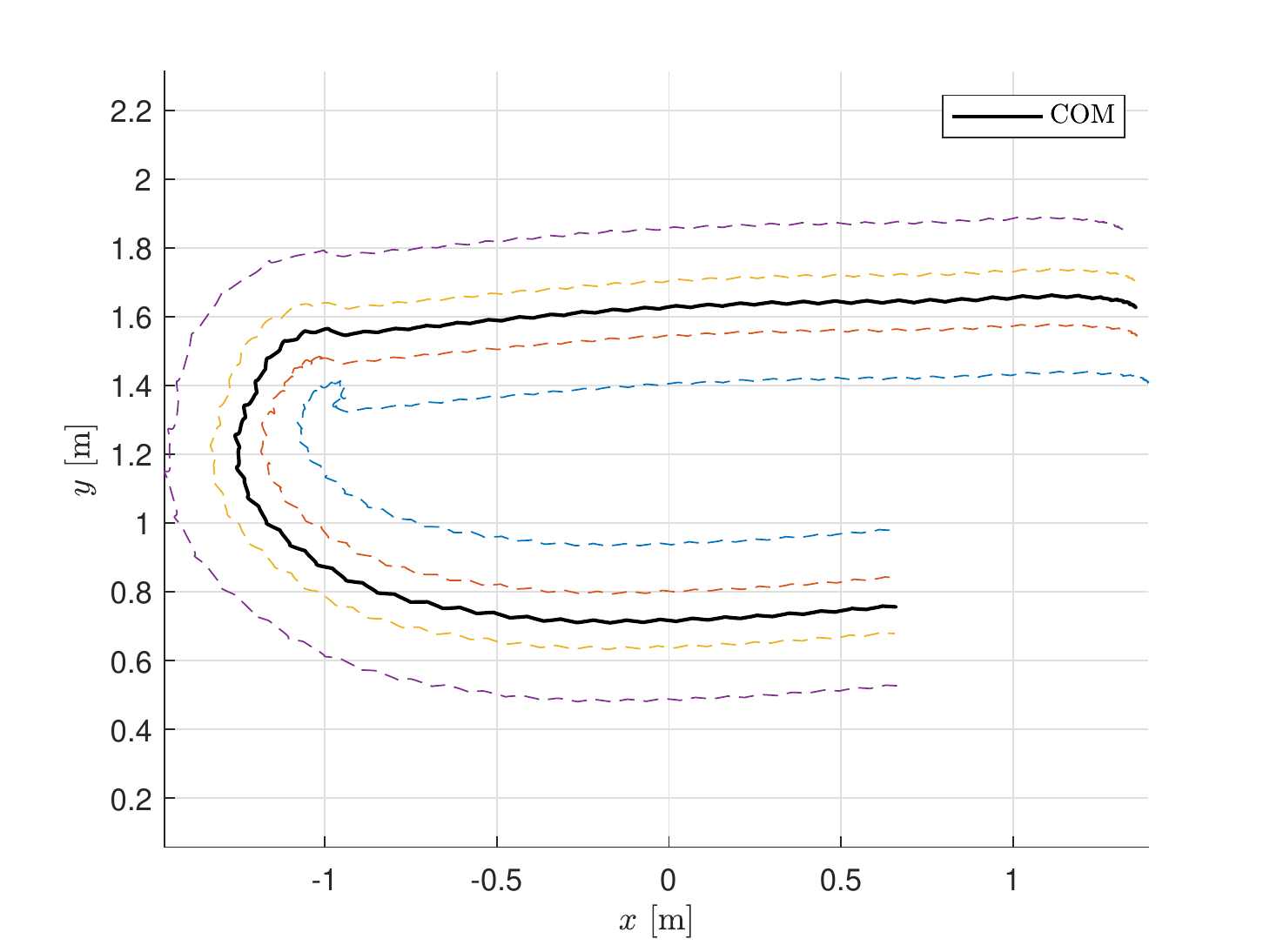}
    \caption{Five boat parallel configuration tracking a yaw of $\pi~\si{rad}$ (left) for $45~\si{s}$, then tracking a yaw of $0~\si{rad}$ (right) for $45~\si{s}$, all while maintaining a velocity of $6~\si{cm/s}$. Each dashed color is an individual module, and the the solid black line is the center of mass.}
    \label{fig:trajectory}
\end{figure}

\begin{figure}[t]
    \centering
    \includegraphics[width = \linewidth, trim={0 0 0 0.5cm}]{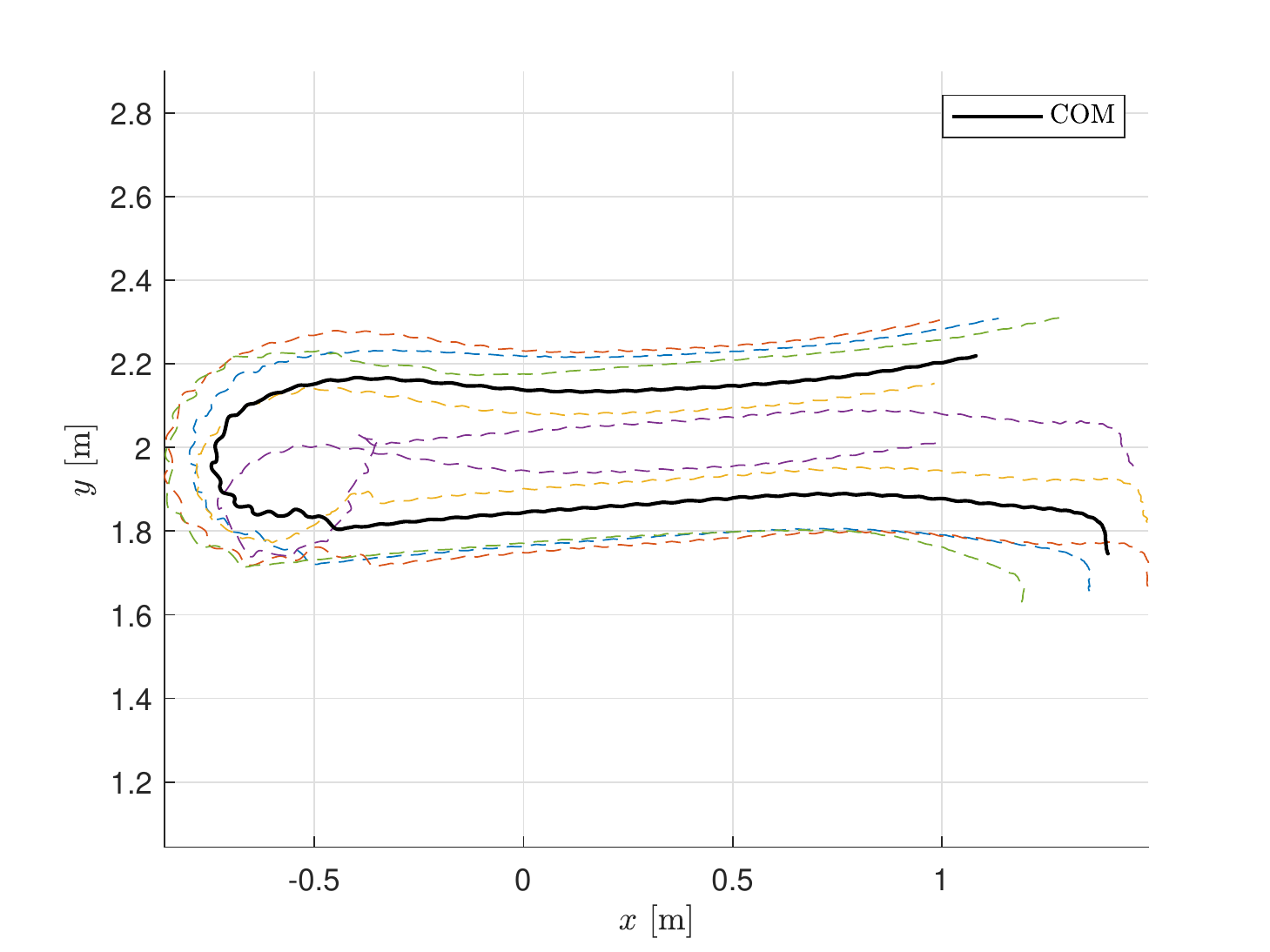}
    \caption{Five boat L configuration tracking a yaw of $\pi~\si{rad}$ (left) for $45~\si{s}$, then tracking a yaw of $0~\si{rad}$ (right) for $45~\si{s}$, all while maintaining a velocity of $4~\si{cm/s}$. Each dashed color is an individual module, and the the solid black line is the center of mass. Note the oscillations during the turn are the result of deformation of the configuration due to fluid forces.}
    \label{fig:trajectoryL}
\end{figure}

The controller presented in Sec.~\ref{sec:dynamics} is capable of tracking a desired surge velocity and yaw angle for an arbitrary configuration. To evaluate its performance, we considered:

\begin{enumerate}
    \item Controlling yaw only for step inputs of either $\pi/2$ or $\pi~\si{rad}$, with the velocity controller deactivated. Representative results for parallel configurations are shown in Fig.~\ref{fig:velResultsP}.
    \item Controlling velocity only for desired velocities of either $4$ or $6~\si{cm/s}$, with the yaw controller deactivated. Representative results for parallel configurations are shown in Fig.~\ref{fig:yawResultsP}.
    \item Controlling both yaw and velocity simultaneously. For these tests, the configurations swam at either $4$ or $6~\si{cm/s}$ at a prescribed yaw angle for $45-60~\si{s}$ and were then given a yaw step input of $\pi~\si{rad}$ while the desired velocity was held constant. Two sample such trajectory are shown in \cref{fig:trajectory,fig:trajectoryL}.
\end{enumerate}

Testing was performed on four parallel configurations, from two to five boats in a parallel line (i.e. along the $\hat{x}_S$ axis), and on five non-parallel configurations, which are shown in Fig.~\ref{fig:diagramsForExperiments}. The results are summarized in \cref{tab:resultsRMS,tab:resultsYawRise,tab:resultsVelRise}. We also considered how the system would behave under mismatched drag coefficients, since it is impossible to experimentally evaluate the drag coefficients for each configuration. To consider the most extreme example of this mismatch, the five boat parallel configuration was also tested with artificially assigned drag coefficients for the two-boat parallel configuration. Representative results for these evaluations are shown in Fig.~\ref{fig:mismatched}.


\section{Discussion} \label{sec:discussion}

The controller presented in this work is intended to provide a general formulation for control of surge velocity and yaw for arbitrary configurations of Modboats. Thus, we would expect that --- for a successful controller --- performance is consistent and independent of the particular configuration. The experimental results presented in Sec.~\ref{sec:experiments} show that this is the case: arbitrary configurations of Modboats can be driven at a desired surge velocity and to a given yaw angle. With a suitable outer control law, this performance can easily be extended to waypoint tracking and more complex behaviors. 

\begin{figure}[t]
    \centering
    \includegraphics[width=\linewidth]{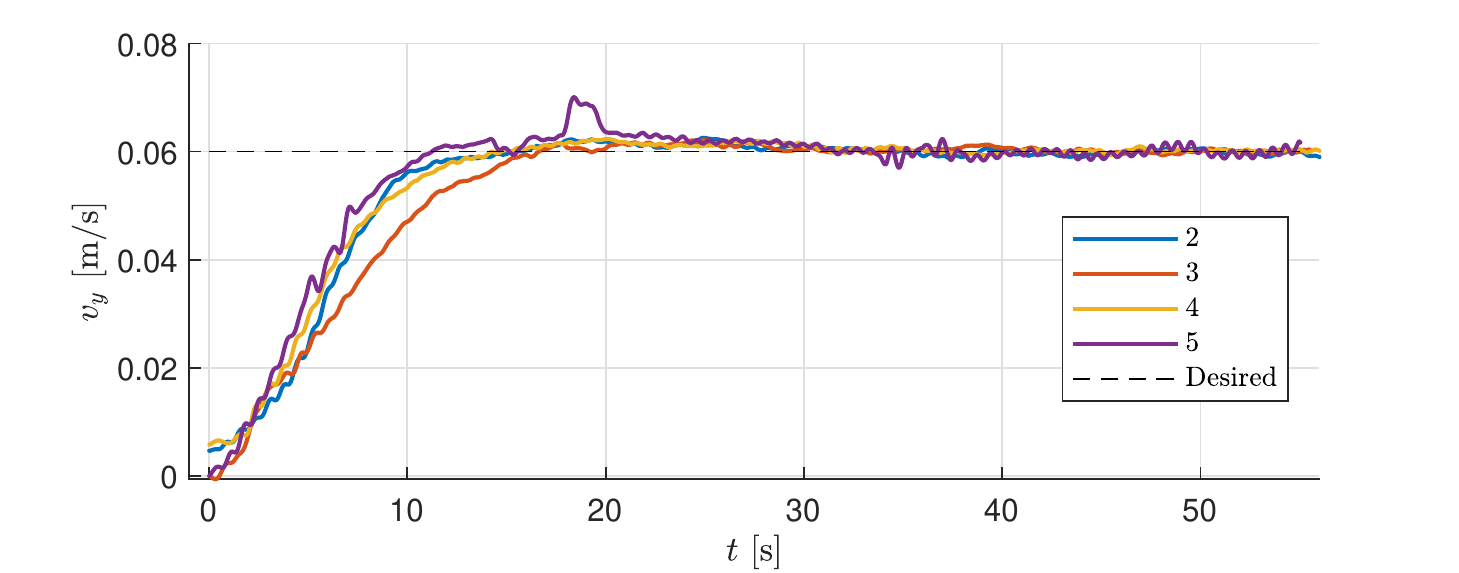}
    \caption{Velocity response to a step input of $6~\si{cm/s}$ for configurations of 2-5 boats in parallel. The controller provides comparable performance regardless of the number of boats in the configuration.}
    \label{fig:velResultsP}
\end{figure}

Velocity tracking, presented in Fig.~\ref{fig:velResultsP} for parallel configurations and summarized in \cref{tab:resultsRMS}, is highly effective regardless of the configuration. All of the evaluated configurations were able to track the desired velocity to within $0.33~\si{cm/s}$ despite varying shapes and interfaces with the surrounding fluid, and non-parallel configurations have statistically identical performance to parallel configurations, indicating that our simple thrust-loss model is sufficient for good performance in velocity tracking.

\begin{table}[t]
    \centering 
    \caption{Controller \textbf{Yaw} rise time (0.3--0.9 of the step) performance from yaw only tests as IQR.
    Parallel configurations are summarized as \textbf{Par}. Non-parallel configurations are summarized as \textbf{NP}.}
    \setlength{\tabcolsep}{5.0pt}
    \begin{tabular}{c|c|c|c}
    \toprule
        ~ & $\pi/2~\si{rad}$ step $[\si{s}]$ &  $\pi~\si{rad}$ step $[\si{s}]$ & Overall $[\si{s}]$ \\ \midrule
        2 & $[2.9,3.8]$* & $[3.0,3.4]$* & $[2.9,3.6]\hphantom{*}$ \\
        3 & $[3.7,3.8]$* & $[4.4,4.5]$* & $[3.7,4.5]\hphantom{*}$ \\
        4 & $[4.9,5.4]$* & $[6.1,6.6]$* & $[5.1,6.4]\hphantom{*}$ \\
        5 & $[4.9,5.1]$* & $[7.3,7.7]$* & $[5.0,7.5]\hphantom{*}$ \\ \midrule
       \textbf{Par.} & $[3.7, 5.0]\hphantom{^\dagger}$ & $[3.9, 6.9]\hphantom{*}$ & $[3.7, 5.8]\hphantom{^\dagger}$ \\ 
        \textbf{NP} & $[2.7, 3.3]^\dagger$ & $[4.1, 5.1]\hphantom{*}$ & $[2.9, 4.4]^\dagger$ \\ \midrule 
        Square & $[2.3, 2.5]\hphantom{*}$ & $[3.2, 3.7]\hphantom{*}$ & $[2.4, 3.4]\hphantom{*}$ \\
        Tetr. & $[3.3, 3.3]\hphantom{*}$ & $[4.1, 4.6]\hphantom{*}$ & $[3.3, 4.2]\hphantom{*}$ \\
        P & $[2.9, 3.6]\hphantom{*}$ & $[4.2, 4.5]\hphantom{*}$ & $[3.1, 4.3]\hphantom{*}$ \\ 
        U & $[2.8, 3.1]\hphantom{*}$ & $[4.4, 5.0]\hphantom{*}$ & $[2.9, 4.6]\hphantom{*}$ \\ 
        L & $[2.8, 3.4]\hphantom{*}$ & $[5.5, 5.7]\hphantom{*}$ & $[3.1, 5.6]\hphantom{*}$ \\ \bottomrule
        \multicolumn{4}{l}{$*$ Data from two test repetitions, so not strictly IQR.} \\
        \multicolumn{4}{l}{$^\dagger$ Statically significant difference over the other summary case} \\
        \multicolumn{4}{l}{in each column.} 
    \end{tabular}
    \label{tab:resultsYawRise}
\end{table}

\begin{table}[t]
    \centering 
    \caption{Controller \textbf{Velocity} rise time (0.3--0.9 of the step) performance from velocity only tests as IQR.
    Parallel configurations are summarized as \textbf{Par}. Non-parallel configurations are summarized as \textbf{NP}.}
    \setlength{\tabcolsep}{5.0pt}
    \begin{tabular}{c|c|c|c}
    \toprule
        ~ & $4~\si{cm/s}$ step $[\si{s}]$ &  $6~\si{cm/s}$ step $[\si{s}]$ & Overall $[\si{s}]$ \\ \midrule
        2 & $[3.8, 12.6]$* & $[4.1,\hphantom{0}4.9]$* & $[4.0, \hphantom{0}8.7]\hphantom{*}$\\
        3 & $[3.7, \hphantom{0}5.1]$* & $[4.4,\hphantom{0}8.7]$* & $[4.1,\hphantom{0}6.9]\hphantom{*}$ \\
        4 & $[5.8, \hphantom{0}6.5]$* & $[4.8,\hphantom{0}6.1]$* & $[5.3,\hphantom{0}6.3]\hphantom{*}$ \\
        5 & $[5.2, \hphantom{0}7.8]$* & $[5.6,\hphantom{0}6.2]$* & $[5.4,\hphantom{0}7.0]\hphantom{*}$ \\ \midrule
       \textbf{Par.} & $[4.4, \hphantom{0}7.2]\hphantom{*}$ & $[4.6, \hphantom{0}6.1]\hphantom{^\dagger}$ & $[4.6, \hphantom{0}6.4]\hphantom{^\dagger}$ \\ 
        \textbf{NP} & $[5.8, \hphantom{0}7.2]\hphantom{*}$ & $[7.5, \hphantom{0}9.2]^\dagger$ & $[6.2, \hphantom{0}8.5]^\dagger$ \\ \midrule 
        Square & $[7.1, \hphantom{0}8.7]\hphantom{*}$ & $[9.3, 12.1]\hphantom{*}$ & $[7.5, 10.5]\hphantom{*}$ \\
        Tetr. & $[5.3, \hphantom{0}5.8]\hphantom{*}$ & $[6.8, \hphantom{0}8.3]\hphantom{*}$ & $[5.5, \hphantom{0}7.7]\hphantom{*}$ \\
        P & $[5.6, \hphantom{0}6.0]\hphantom{*}$ & $[7.3, \hphantom{0}8.4]\hphantom{*}$ & $[5.8, \hphantom{0}7.8]\hphantom{*}$ \\ 
        U & $[6.0, \hphantom{0}6.3]\hphantom{*}$ & $[7.7, \hphantom{0}9.1]\hphantom{*}$ & $[6.1, \hphantom{0}8.4]\hphantom{*}$ \\ 
        L & $[6.9, \hphantom{0}7.6]\hphantom{*}$ & $[7.8, \hphantom{0}8.9]\hphantom{*}$ & $[7.2, \hphantom{0}8.3]\hphantom{*}$ \\ \bottomrule
        \multicolumn{4}{l}{$*$ Data from two test repetitions, so not strictly IQR.} \\
        \multicolumn{4}{l}{$^\dagger$ Statically significant difference over the other summary case} \\
        \multicolumn{4}{l}{in each column.}
    \end{tabular}
    \label{tab:resultsVelRise}
\end{table}

\begin{figure}[t]
    \centering
    \includegraphics[width=\linewidth]{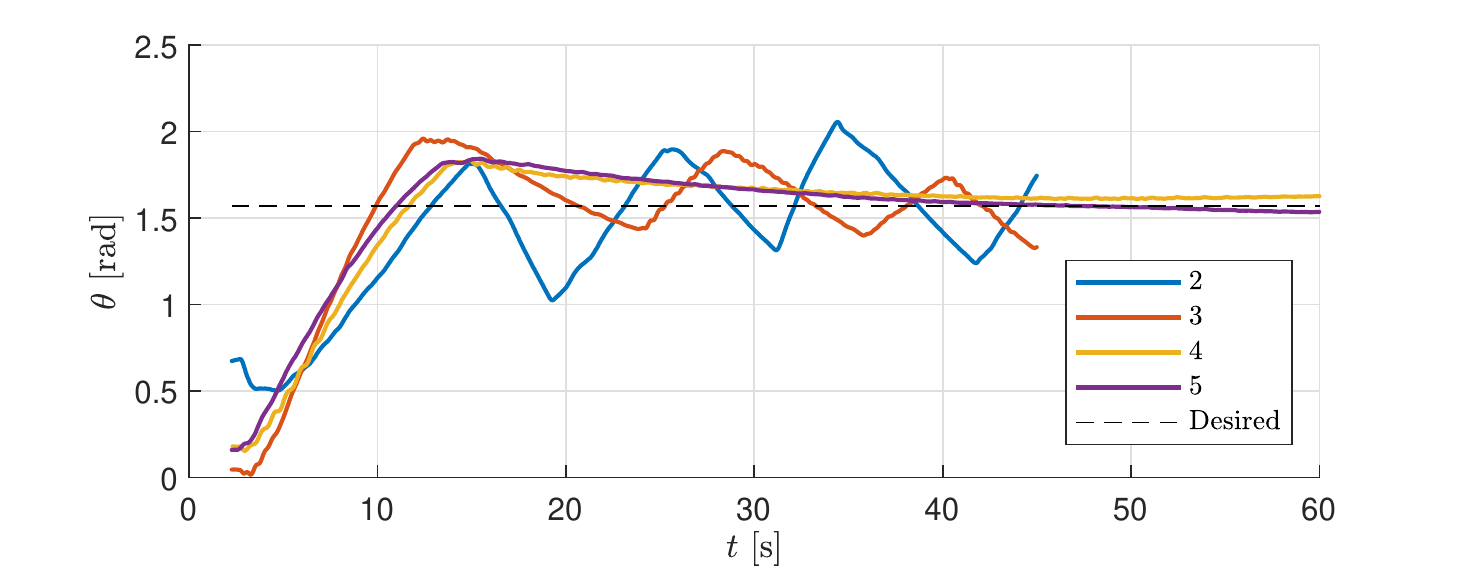}
    \caption{Yaw response to a step input of $\pi/2~\si{rad}$ for configurations of 2-5 boats in parallel. The controller provides comparable rise time performance regardless of the number of boats in the configuration, but oscillation amplitudes decrease as the configuration grows.}
    \label{fig:yawResultsP}
\end{figure}

\begin{figure}[t]
    \centering
    \subfloat[\label{fig:mismatchedVel}]{\includegraphics[width=0.98\linewidth, trim={0.5cm 0 0.5cm 0.5cm}, clip]{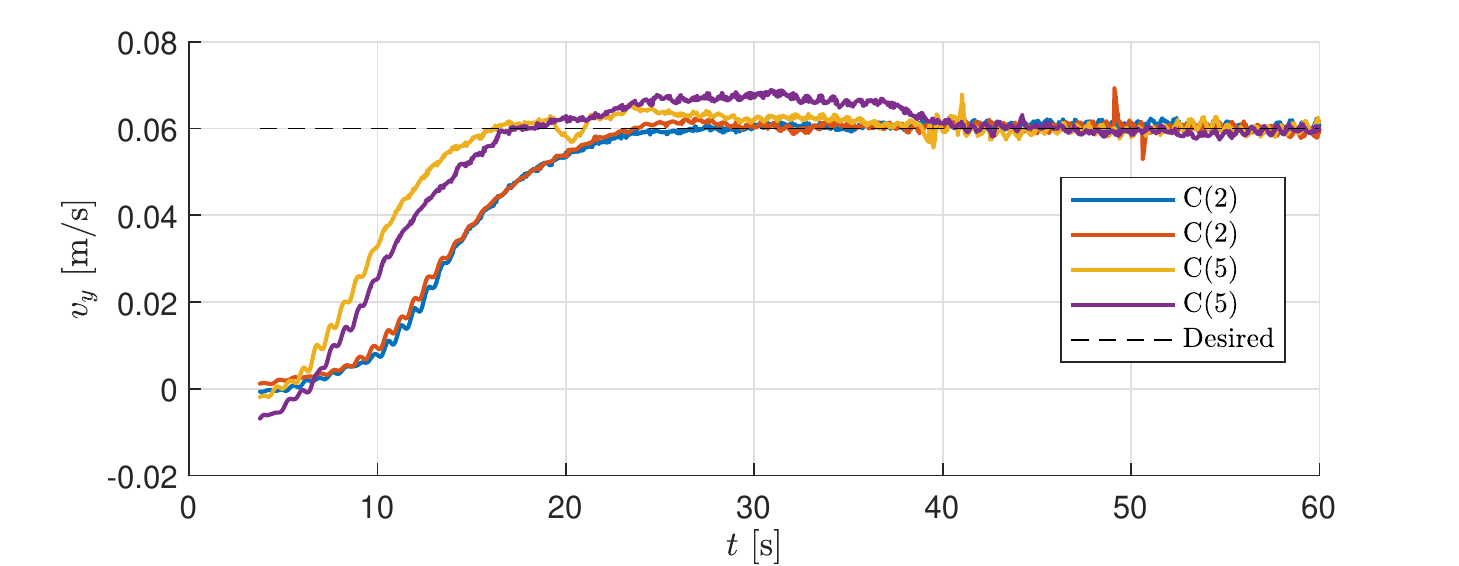}
      }
          \hfill
    \subfloat[\label{fig:mismatchedYaw}]{\includegraphics[width=0.98\linewidth, trim={0.5cm 0 0.5cm 0.5cm}, clip]{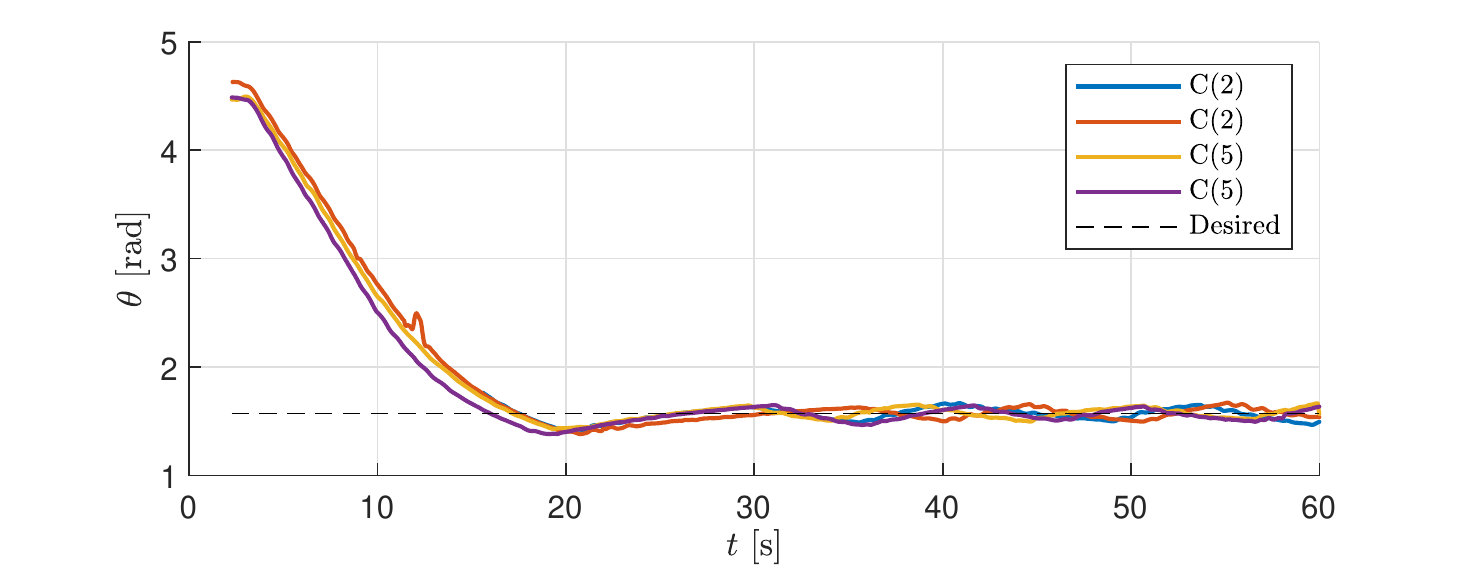}
        }
    \caption{(a) Velocity and (b) yaw step response for the $5$ boat configuration for where the drag coefficients are chosen as $C_L(n)$ and $C_R(n)$ with $n=5$ and $n=2$. Two tests are shown for each.}
    \label{fig:mismatched}
\end{figure}

Yaw tracking, presented in Fig.~\ref{fig:yawResultsP} for parallel configurations and summarized in \cref{tab:resultsRMS}, is similarly agnostic to the particular configuration. Yaw tracking is excellent when swimming, as small adjustments to the amplitude of each boat's oscillation yaw the configuration and track the desired heading to within $0.11~\si{rad}$ ($6.5^\circ$). Yaw tracking suffers, however, when no velocity is desired; although the yaw can be driven to the desired value overall, the tracking error increases significantly. This occurs because the configuration is attempting to remain in place, but finer adjustments require reversing thrust. If there is enough initial overshoot, significant oscillations around the desired value begin and are not damped out by the derivative gain, such as for the two and three boat configurations in Fig.~\ref{fig:yawResultsP} for a $\pi/2~\si{rad}$ step input and for the four and five boat configurations in Fig.~\ref{fig:mismatchedYaw} for a $\pi~\si{rad}$ step input. Otherwise, a significant settling time is observed as the boats attempt to overcome the large inertia of the configuration, such as for the four and five boat configurations in Fig.~\ref{fig:yawResultsP}.

\begin{figure}[t]
    \centering
    \includegraphics[width=\linewidth, trim={0 0 0 0.5cm}]{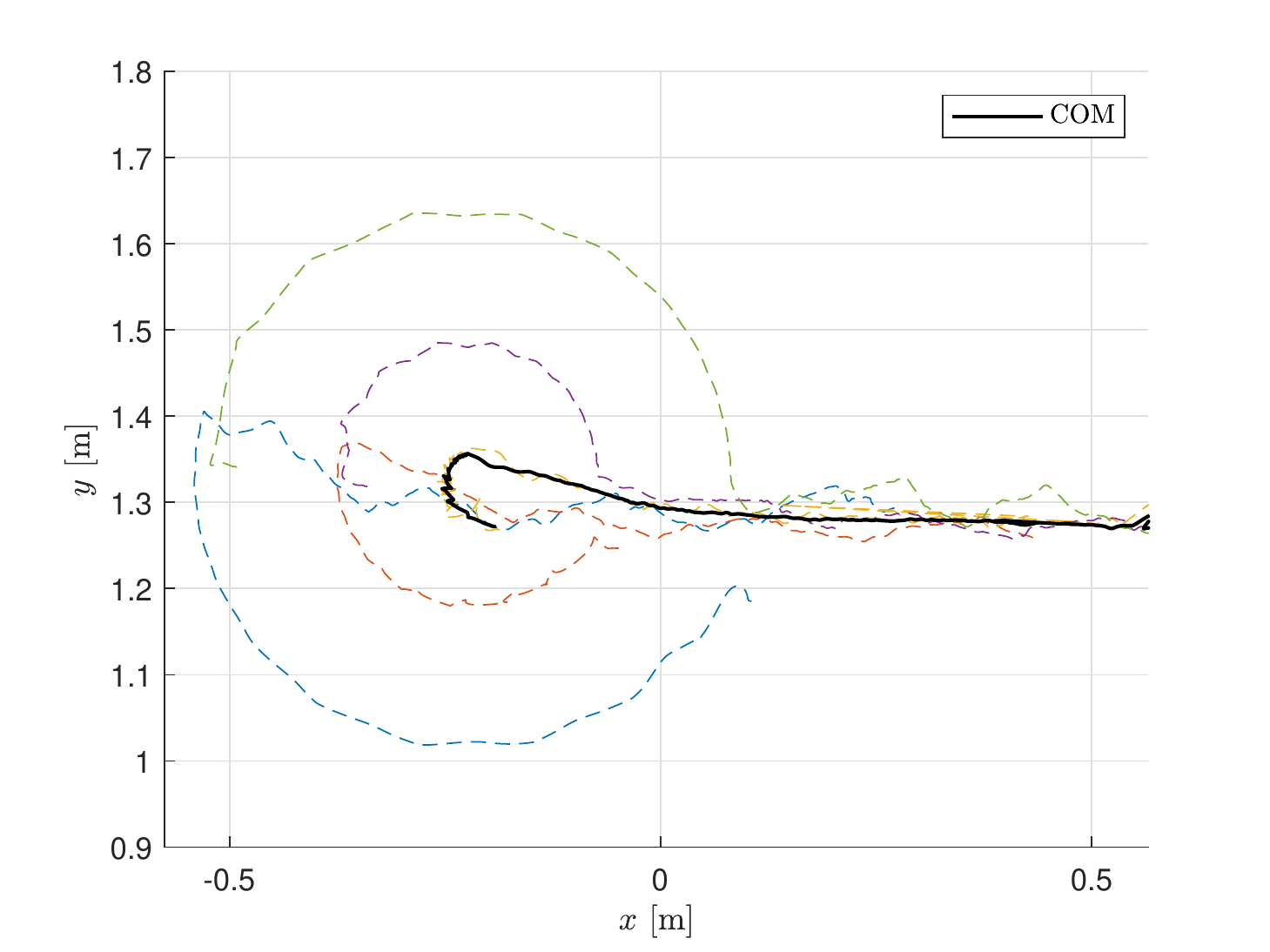}
    \caption{Trajectory of five boat configuration responding to a $\pi~\si{rad}$ step input without a desired velocity. Dotted lines show individual boat trajectories, while the solid line follows the center of mass. The configuration turns in place until the desired yaw is achieved, and then drifts uncontrollably sideways due to the discontinuity in~\eqref{eq:waveformNew}.}
    \label{fig:sidewaysAfterYaw}
\end{figure}

A secondary problem with controlling yaw alone is shown in Fig.~\ref{fig:sidewaysAfterYaw}; after achieving the desired yaw with minimal motion of the center of mass (COM), the configuration drifts uncontrollably along its $x$ axis. Since our controller can produce forces only along its surge axis, we have no way of counteracting this behavior. Both issues stem from the same cause: the Modboats' unique propulsive mechanism cannot smoothly transition from forward to reverse. Eq.~\eqref{eq:waveformNew} is strongly discontinuous when $(\phi_0)_i$ changes, especially if $A_i$ is small\footnote{Note that~\eqref{eq:waveformNew} is discontinuous whenever $A_i$ changes even if $(\phi_0)_i$ is constant. We do not observe significant effects from this, however, and it is the normal mode of operation for the control strategy presented in this work.}. The transition from forward to reverse thrust thus creates sideways forces and yaw torques that disrupt the controller.

Nevertheless, Table~\ref{tab:resultsRMS} shows that the yaw tracking performance is remarkably consistent across configuration shapes, and non-parallel configurations have statistically identical performance to parallel configurations, indicating that our simple thrust-loss model is sufficient for good performance in yaw tracking. 

More significant distinctions emerge when considering the rise-time achieved by the controller. Fig.~\ref{fig:mismatchedVel} shows that while using the wrong drag coefficient results in comparable tracking performance, it does result in worse velocity rise time. This is reasonable, since drag is the only term in the velocity equation of~\eqref{eq:forces}. Similarly, Table~\ref{tab:resultsVelRise} shows that non-parallel configurations have rise-times that are generally worse than those of parallel configurations. Drag coefficients for non-parallel configurations are determined through projection and underestimated (see Sec.~\ref{sec:drag} and Fig.~\ref{fig:dragCoeffs}), so it is reasonable that longer rise-times result.

Rise-times for yaw do not show this same pattern, and non-parallel configurations show a small \textit{improvement} over parallel configuration behavior, as shown in Table~\ref{tab:resultsYawRise}. This is consistent with the behavior observed when the drag is mismatched, as in Fig.~\ref{fig:mismatchedYaw}, where the performance is not significantly affected. Having no significant shift in performance is reasonable, since in the yaw equation of~\eqref{eq:forces} we observe that $I\alpha \gg C_R\abs{\Omega}{\Omega}$. The inertia term --- which is calculated far more accurately --- dominates over the effect of the less well modeled drag. 

\section{Conclusion} \label{sec:conclusion}

In this work we have presented a \textbf{centralized control approach} that allows \textbf{arbitrary configurations of underactuated modules} to \textbf{swim as a single unit}, as long as they can produce thrust along a single axis. When applied to configurations of Modboats, this strategy minimizes internal forces, \textbf{guarantees no undocking}, and allows the non-rigid configuration to behave similarly to a rigid body. Using a minimal model of fluid interactions between neighboring modules and a small set of drag coefficient measurements, our  controller is capable of effectively tracking a desired surge velocity and yaw angle for a variety of differently shaped configurations. These results have been verified experimentally for configurations consisting of two to five boats, in nine different configurations.

Velocity tracking is shown to be highly effective, and yaw tracking while swimming forward is similarly accurate. The controller struggles to track yaw while stationary however, managing to get within the desired region but generating oscillations and sideways drift that the control law cannot counteract. Thus this controller is poorly suited for docking or station keeping, both of which require precise orientation control while stationary, but is well suited for transportation of objects or collective travel. Future work will consider ways to extend the control law to the configuration's sway axis and reduce the observed oscillations in yaw.

A number of approaches for controlling smaller configurations --- i.e. single Modboats --- already exist~\cite{Knizhnik2020a,Knizhnik2021a}. For larger configurations we theorize that our controller will continue to perform well, although problems may arise as the yaw authority of individual boats scales linearly with their distance from the center of mass, but the angular drag and inertia scale quadratically. An additional issue to consider is the wake compensation factor and the maximum allowable amplitude causing increased clipping as the configurations grow vertically. Velocity tracking is likely to remain effective, but our testing tank is too small for larger configurations or extensive maneuvering.

Finally, although the current control formulation is sufficient to guarantee no unintentional collisions, it is not necessary. In fact, for $N \geq 3$ boats it should be possible to \textit{simultaneously} control both translation axes and rotation, rendering any arbitrary configuration into a holonomic vehicle despite the limitations of each module. Finding a control law to accomplish this, however, would require dynamic solutions that avoid the complex collision space within an arbitrary configuration. Future work will investigate such control laws.


\section*{Acknowledgment}

We thank Dr. M. Ani Hsieh for the use of her instrumented water basin in obtaining all of the testing data, and Peihan M. Li and Julia Dase for their help running experiments for this work. 

\bibliographystyle{./bibliography/IEEEtran}
\bibliography{./bibliography/IEEEabrv,./bibliography/nonpaper,./bibliography/references}



\appendices


\section{Proof of Identical Forces for Vertical Neighbors} \label{app:sameForces}

Our proof for avoiding collisions between vertical neighbors in Sec.~\ref{sec:selfCollisionNP} assumes that vertical neighbors make the same choice of centerline $\phi_0$, since non-adherence to this restrictions results in collisions in many cases. In fact, this is a mathematical consequence of the use of the matrix $P$ and its pseudo-inverse, since boats with identical values of $x_i$ (which includes all vertical neighbors) will have the same force output. 

Consider the structural matrix $P$ parameterized by its columns given in~\eqref{eq:structuralMatrixCols}. Note that, as per~\eqref{eq:structuralMatrix}, each column has the form $p_i = [\begin{matrix} 1 & x_i \end{matrix}]^T$.

\begin{equation} \label{eq:structuralMatrixCols}
    P = \begin{bmatrix} p_1 & \hdots & p_N \end{bmatrix}
\end{equation}

In the pseudo-inverse, let $A = (P P^T)^{-1} \in \mathbb{R}^{(2\times 2)}$. Then~\eqref{eq:pseudoInverseRows} shows the form of the pseudo-inverse $P^+$.

\begin{equation} \label{eq:pseudoInverseRows}
    P^+ = P^T(P P^T)^{-1} = P^T A = \begin{bmatrix} p_1^T \\ \vdots \\ p_N^T \end{bmatrix} A = \begin{bmatrix} p_1^T A \\ \vdots \\ p_N^T A \end{bmatrix}
\end{equation}

We see from~\eqref{eq:pseudoInverseRows} that if we have two columns $i$ and $j$ of $P$ that are identical, such that $p_i = p_j$ (i.e. boats in the same vertical column) then the corresponding rows $i$ and $j$ of $P^+$ will also be identical.

\begin{equation}\label{eq:forceAssignmentRows}
\vec{f} = \begin{bmatrix} f_1 \\ \vdots \\ f_N \end{bmatrix} =  P^+ \vec{F} = \begin{bmatrix} p_1^T A \cdot \vec{F} \\  \vdots \\ p_N^T A \cdot \vec{F} \end{bmatrix} 
\end{equation}

Eq.~\eqref{eq:forceAssignmentRows} translates from the configuration frame forces $\vec{F}$ to the individual boat forces $f_i$, and we can see that if $x_i = x_j$, then $p_i = p_j$ and $f_i = f_j$. Since the centerline $\phi_0$ is determined by $\sign{(\phi_0)}$, $(\phi_0)_i = (\phi_0)_j$. \qed

One final challenge is provided by the hydrodynamic model in Sec.~\ref{sec:hydrodynamics}, since $f_i$ is artificially inflated by a factor of $1/\alpha_i$ before being used. However, since $\alpha_i > 0$, it is always true that $\sign{(f_i/\alpha_i)} = \sign{(f_i)}$, so the validity of the proof is unaffected. The potential for collision provided by the increased amplitude is addressed in Appendix~\ref{app:verticalNeighbors}.

\vspace{0.25cm}


\section{Proof of Non-Collision Despite Amplitude Increase} \label{app:verticalNeighbors}

The Modboat tail is parameterized in $(r,\theta)$ by~\eqref{eq:tailFunc}, where $r_t$ is the radius of the top body, $r_p$ is the protrusion of the tail tip, and $\phi$ is the tail angle (values given in Table~\ref{tab:tailParams} for the Modboats used in this work). The subscript $(-\pi,\pi]$ indicates wrapping to that range, and $\theta_w$ is a parameter setting the width of the protruding region. Eq.~\eqref{eq:tailFunc} causes the radius to increase linearly in proportion to angle. This allows two boats to smoothly increase the distance between them when undocking.
\begin{equation} \label{eq:tailFunc}
    r(\theta) = r_t + r_p \left ( 1 - \abs{\frac{(\phi-\theta)_{(-\pi,\pi]}}{\theta_w}} \right )
\end{equation}

\begin{figure}[t]
    \centering
    \includegraphics[trim={2.0cm 0 2.0cm 1.00cm},clip, width=\linewidth]{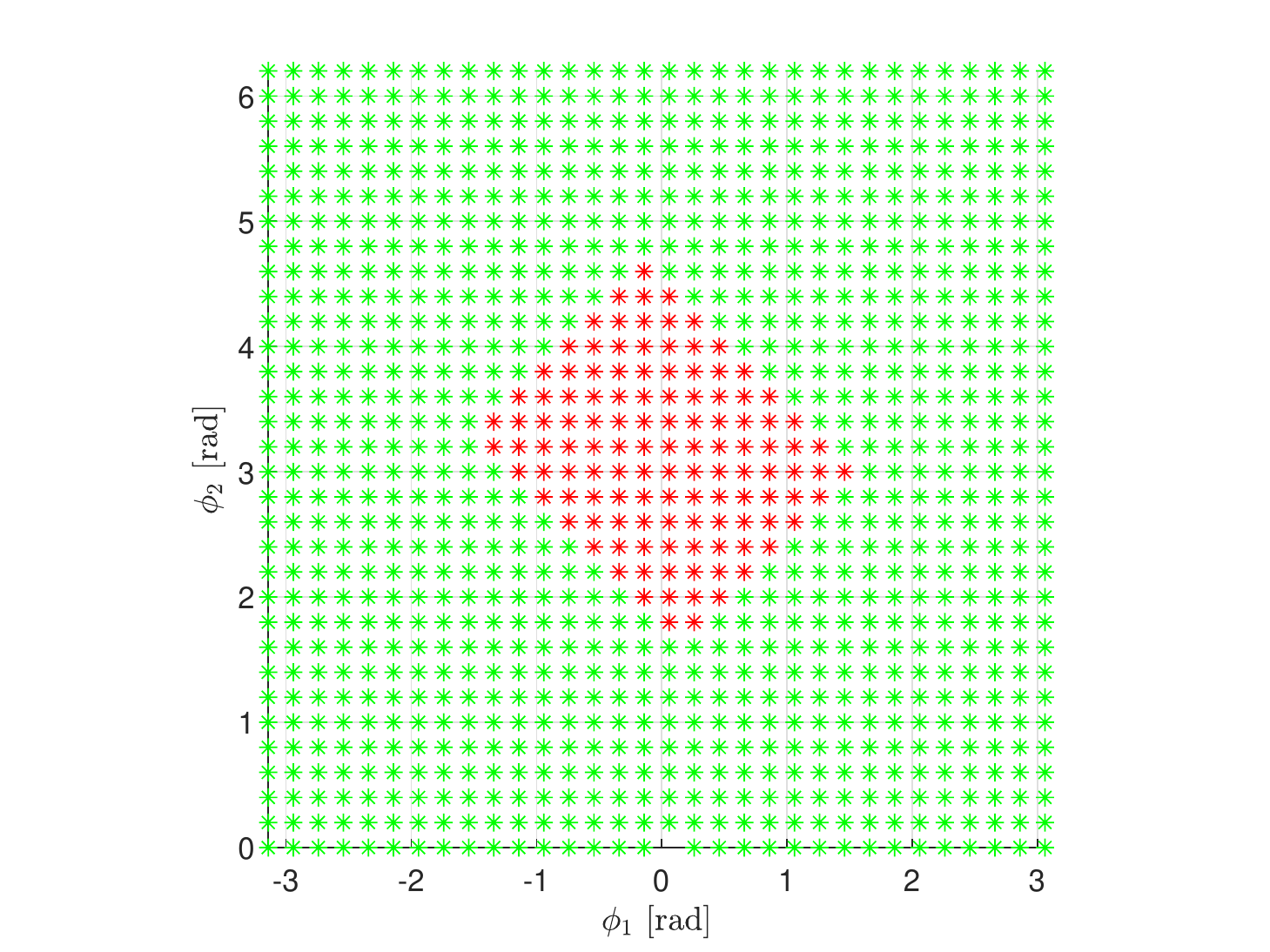}
    \caption{Numerical model of collision space between horizontal neighbors with boat $2$ behind boat $1$. The green region represents no collision, while the red diamond marks the collision space.}
    \label{fig:collisionSpace}
\end{figure}

We can numerically model the entire collision space for two neighboring boats, which results from intersections between two tails modeled by~\eqref{eq:tailFunc} separated by $2r_t$, and the resulting space is shown in Fig.~\ref{fig:collisionSpace} for a front-back neighbor pair (recall that $\phi_i=0$ has the tail tip pointing rearward, and $\phi_i=\pi$ has the tail tip pointing forward).

We consider Modboats executing waveforms of the form given in~\eqref{eq:waveformNew}. Assumption~\ref{asp:phaseLock} guarantees that $\omega$ is the same for both boats, and since we consider only vertical neighbors we know that both must make the same choice of centerline $\phi_0$ as a consequence of the matrix inverse in~\eqref{eq:forces}. Consider also that --- as dictated by our approach --- boats 1 and 2 have their amplitude related as in~\eqref{eq:gamma}.

Then we have:

\begin{align}
    \frac{d\phi_2}{d\phi_1} &= \frac{d\phi_2/dt}{d\phi_1/dt} \\
        &= \frac{-A_{2} \sin{(\omega t)}\cos{\left ((\phi_0)_{2} \right )}}{ - A_{1} \sin{(\omega t)}\cos{\left ((\phi_0)_{1} \right )}} \\
        &= \frac{A_2}{A_1} = \gamma(2) \label{eq:slopeIsGamma}
\end{align}

Eq.~\ref{eq:slopeIsGamma} shows that trajectories in the phase space shown in Fig.~\ref{fig:collisionSpace} are \textit{lines} with slope dictated entirely by $\gamma$. Moreover, because of Assumption~\ref{asp:phaseLock} all trajectories must pass through the point $(0,0)$. 

Thus we can consider the effect of $\gamma$ by drawing a horizontal line passing through the origin and slowly increasing the slope until a collision occurs at $\gamma_{max} \approx 1.9$.

\begin{table}[t]
    \centering
    \caption{Tail parameters used in~\eqref{eq:tailFunc}.}
    \begin{tabular}{c|ccc} \toprule
    Parameter & $r_t~[\si{m}]$ & $r_p~[\si{m}]$ & $\theta_w~[\si{rad/deg}]$ \\ \midrule
    Value  & $0.0762$ & $0.015$ & $0.62/35$  \\ \bottomrule
    \end{tabular}
    \label{tab:tailParams}
\end{table}

Since all $\gamma$ values provided in Table~\ref{tab:parameters} are below $\gamma_{max}$, no collisions will occur. \qed

\clearpage

\end{document}